\documentclass[10pt,twocolumn,letterpaper]{article}

\usepackage{cvpr}
\usepackage{times}
\usepackage{epsfig}
\usepackage{graphicx}
\usepackage{array}
\usepackage{tabularx}
\newcolumntype{L}{>{\raggedright\arraybackslash}X}
\usepackage{caption, booktabs}
\usepackage{amsmath}
\usepackage{amssymb}
\usepackage{float}
\usepackage{placeins}
\usepackage{bm}
\DeclareMathOperator*{\argmax}{arg\,max}

\usepackage[pagebackref=true,breaklinks=true,letterpaper=true,colorlinks,bookmarks=false]{hyperref}

\usepackage{hhline}

\cvprfinalcopy 

\def\cvprPaperID{****} 

\pdfoutput=1
\begin{document}


\def\eqnvspace{{\vspace{-2mm}}}
\def\figvspace{{\vspace{-4mm}}}

\newcommand{\Paragraph}[1]{\vspace{1.25mm} \noindent \textbf{#1} \hspace{0mm}}
\newcommand{\Section}[1]{\vspace{-1.5mm} \section{#1} \vspace{-1.5mm}}
\newcommand{\SubSection}[1]{\vspace{-1.2mm} \subsection{#1} \vspace{-1.2mm}}
\newcommand{\SubSubSection}[1]{\vspace{-3mm} \subsubsection{#1} \vspace{-1mm}}

\newcommand{\norm}[1]{\left\lVert#1\right\rVert}

\title{Depth Coefficients for Depth Completion}

\author{Saif Imran \hspace{1cm} Yunfei Long \hspace{1cm}  Xiaoming Liu \hspace{1cm}Daniel Morris \\
Michigan State University\\
428 S. Shaw Ln, EB 2120, East Lansing, MI 48824\\
{\tt\small \{imransai,longyunf,liuxm,dmorris\}@msu.edu}
}

\maketitle

\begin{abstract}

Depth completion involves estimating a dense depth image from sparse depth measurements, often guided by a color image.  While linear upsampling is straight forward, it results in artifacts including depth pixels being interpolated in empty space across discontinuities between objects.  Current methods use deep networks to upsample and "complete" the missing depth pixels.  Nevertheless, depth smearing between objects remains a challenge.  We propose a new representation for depth called Depth Coefficients (DC) to address this problem.  It enables convolutions to more easily avoid inter-object depth mixing. We also show that the standard Mean Squared Error (MSE) loss function can promote depth mixing, and thus propose instead to use cross-entropy loss for DC. With quantitative and qualitative evaluation on benchmarks, we show that switching out sparse depth input and MSE loss with our DC representation and cross-entropy loss is a simple way to improve depth completion performance, and reduce pixel depth mixing, which leads to improved depth-based object detection.
\end{abstract}


\Section{Introduction}

Active depth sensing has achieved significant gains in performance and demonstrated its utility in numerous applications over the last two decades.  High-resolution depth estimation contributes towards 3D scene understanding~\cite{schwarz2010lidar}, object detection~\cite{qi2017frustum}, classification and tracking~\cite{Fortin:2015:Lidar}, and object shape estimation~\cite{cui2013algorithms}.  Important sensors include Intel RealSense and Microsoft Kinect $2$ for indoor applications, and Lidars such as the Velodyne VLP-$64$ for longer-range outdoor applications such as automotive safety and autonomy.  While performance in range and resolution are improving, nevertheless the cost for higher-resolution Lidars remains prohibitive for numerous applications. As a result there is significant ongoing effort into improving the resolution, while lowering the cost of 3D sensors~\cite{Uhrig2017THREEDV,park2011Upsampling,jaritz2018sparse}.

Our application goal is  high-resolution shape analysis and object detection using an inexpensive, low-resolution Lidar, complemented with a high-resolution camera. The key step of using a color image to guide depth super-resolution is called {\it depth completion}.  Current state-of-the-art methods~\cite{mal2018sparse,chen2018estimating, ma2018self}, rely on deep convolutional networks and perform well at up-sampling within-object depths.

\begin{figure}[t!]
    \begin{center}
      \begin{tabular}{c}
        \includegraphics[trim=194 0 270 10,clip,width=\linewidth]{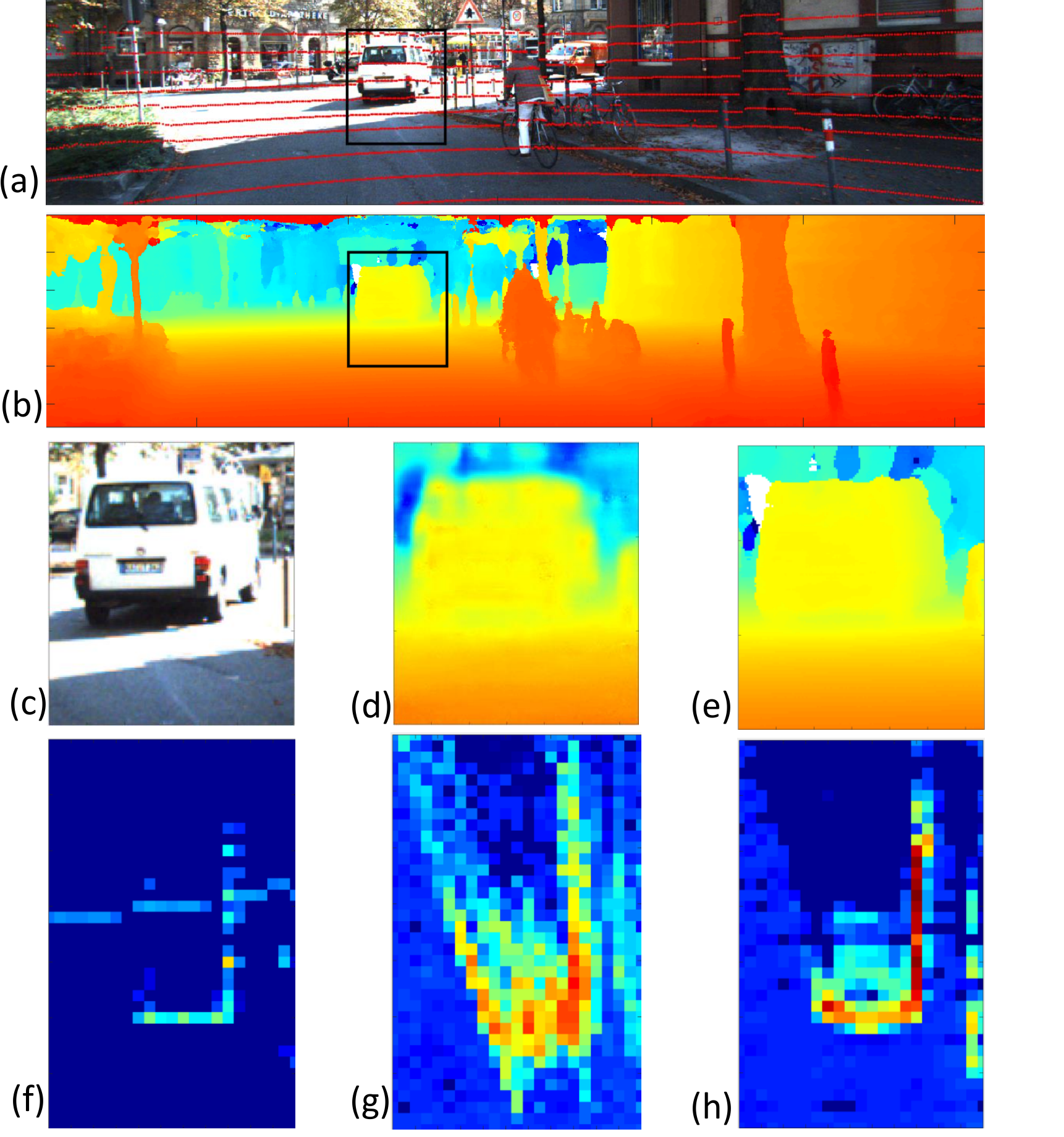} \\
      \end{tabular}
        \vspace{-2mm}
        \end{center}
    \caption{Our depth completion uses ($a$) a color image and the subsampled (16R) Lidar points projected into image plane to estimate ($b$), a dense depth image. ($c$-$e$) are zoomed-in view of input color image, super-resolved depth of Ma et al.~\cite{ma2018self} and ours  respectively. 
    ($f$-$h$) are bird's eye view of input sparse Lidar data, ($d$), and ($e$), respectively.
    Colors in the bird's eye view show the number of height pixels in each cell/pixel. So a smeared object shape has height pixels spread out around the object boundary. Notice the smearing of depth at the object boundaries in ($g$) compared to ($h$). These depth-mixing pixels impact qualitative appearance as well as subsequent tasks, such as object detection and pose estimation.}
    \label{fig:tease_fig}\figvspace
\end{figure}


While these methods score well on Root-Mean-Square Error (RMSE), nevertheless they still generate {\it mixed-depth pixels}. We define mixed-depth pixels as those pixels whose estimated depth places them at neither the foreground nor background object, but in-between the objects. Since mixed-depth pixels occur primarily at depth discontinuities, they typically constitute a small fraction of the total pixel count and various loss measures.  Nevertheless, their impact on the quality of depth maps and projected point-clouds is significant including spurious points in mid-air and connecting surfaces between separate objects.  

This paper aims to investigate the cause of the mixed-depth pixels and propose a solution.  We investigate how current depth-image representation leads to depth mixing, and propose an alternative representation called Depth Coefficients (DC) to avoid this.  We also examine how loss functions, such as MSE, favor mixed-depth pixels in certain cases.  Using our newly proposed DC representation, we leverage cross-entropy loss to avoid promoting depth mixing. Finally we propose a pair of evaluation metrics that can be used in place of, or to complement, RMSE and MAE. Unlike RMSE and MAE, our new metrics penalize mixed-depth pixels and so may be better quality measures for evaluating depth completion.  Sample output is shown in Fig.~\ref{fig:tease_fig}.

The contributions of this paper are: (1) an analysis of the cause of mixed-depth pixels, (2) a novel depth representation that reduces depth mixing, (3) a new use of cross entropy as a depth loss function, (4) two evaluation metrics that penalize mixed-depth pixels, and (5) demonstration of improved object detection from super-resolved depth.

\Section{Related Work}

\Paragraph{Depth completion}
The substantially lower resolution of depth sensors compared to color cameras has been a motivator for depth completion. Early work by Diebel and Thrun~\cite{Diebel2006MRF} used markov random fields to guide upsampling, and this was followed by a variety improvements including bilateral filters~\cite{Yang2007Spatial}, robust regularization~\cite{park2011Upsampling}, hand-crafted filters~\cite{ferstl2013image} and image segmentation~\cite{lu2015sparse}.  More recently deep convolutional neural networks (CNNs) have taken the lead and moved on from the Middlebury dataset~\cite{Scharstein2003} to larger datasets, NYU2~\cite{Silberman:ECCV12} and KITTI~\cite{Geiger2013IJRR}. Leading contenders, including~\cite{mal2018sparse,chen2018estimating, ma2018self}, perform well on these datasets with both regular and irregularly sampled data.  Our work uses a similar network as~\cite{ma2018self}, but with focus on the depth representation and loss function, instead of the architecture.

\Paragraph{Depth representation}
Measurements of $3$D shape can be represented multiple ways, each with its own advantages and drawbacks.  
{\it $3$D point clouds} are widely used in object detection~\cite{qi2017frustum}, segmentation~\cite{brostow2008segmentation} and surface normal estimation~\cite{mitra2003estimating}. Their advantages include being precise, straight from sensors, and that euclidean distances can be calculated between point cloud clusters. However, direct convolutions are not possible with point clouds; object surfaces are not fully represented by point clouds since they are sparse and unorganized. 
{\it Voxels} can provide a regular grid for object detection~\cite{zhou2017voxelnet}, object classification and orientation estimation \cite{riegler2017octnet}, but can be memory intensive at high resolutions.
{\it Depth images}, sometimes considered $2.5$D representations, have been used for RGBD fusion and instance segmentation~\cite{shao2018clusternet,gupta2014learning}.  They naturally encode sensor viewing rays and adjacency between points. They are compact representations and with their regular grids can be processed with CNN in an analogous way to color image super-resolution~\cite{image-super-resolution-via-deep-recursive-residual-network,memnet-a-persistent-memory-network-for-image-restoration}.  This is the representation of choice for colorization techniques and fusion~\cite{Silberman:ECCV12} as well as depth completion. 

While depth images are popular for depth completion, we will examine an important drawback: the tendency to generate mixed-depth pixels between surfaces. One goal of our DC representation is to remedy this drawback. 

\Paragraph{Loss function for depth completion}
A key component of depth completion is the choice of loss function.
Recent work has explored loss functions  including L2~\cite{chen2018estimating,ma2018self}, L1~\cite{mal2018sparse}, inverse-L1~\cite{jaritz2018sparse}, and softmax losses on depth~\cite{liao2017parse}. While these loss functions can achieve low error on measures including RMSE, MAE, iMAE, often it comes at the cost of smoothing out depth estimate at object boundaries. In this way, the sharp boundaries are lost/smeared and object shapes are distorted. We propose to impose cross-entropy on our probabilistic representation, and show this gives both high performance and sharp boundaries.

\Section{Mixed-depth Pixels}
\label{sec:mixing}

\begin{figure}[t!]
    \begin{center}
    \begin{tabular}{@{\hskip1pt}c@{\hskip1pt}c@{\hskip1pt}c@{\hskip1pt}c@{\hskip1pt}c@{\hskip1pt}c@{\hskip1pt}c}
    ($a$) &
    \includegraphics[trim=240 320 260 325,clip,width=0.25\linewidth]{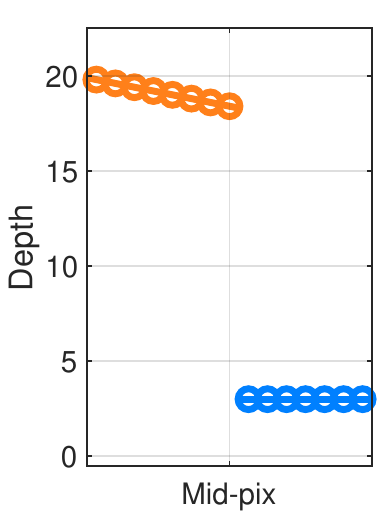} & ($b$) &
    \includegraphics[trim=240 320 260 325,clip,width=0.25\linewidth]{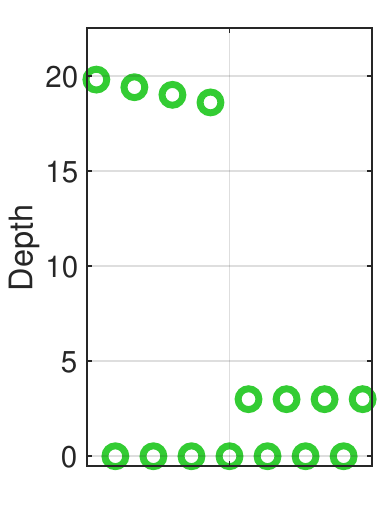} & ($c$) &
    \includegraphics[trim=240 320 260 325,clip,width=0.25\linewidth]{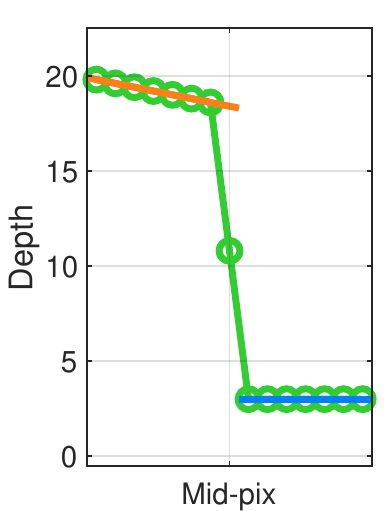}  \vspace{-1mm}\\ 
    ($d$) &
    \includegraphics[trim=240 320 250 325,clip,width=0.25\linewidth]{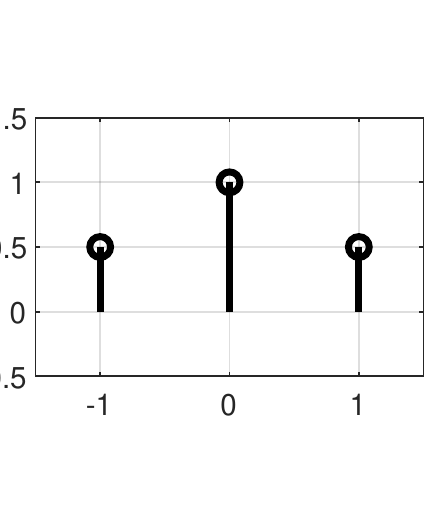} & ($e$) &
    \includegraphics[trim=240 320 260 325,clip,width=0.25\linewidth]{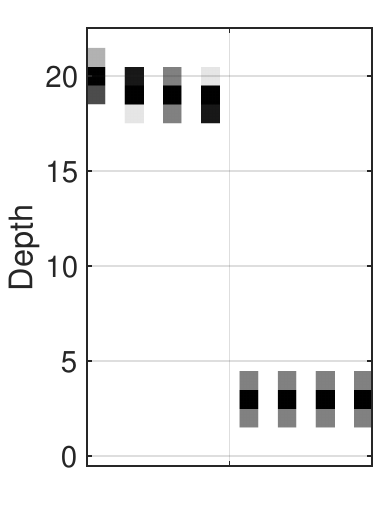} & ($f$) &
    \includegraphics[trim=240 320 260 325,clip,width=0.25\linewidth]{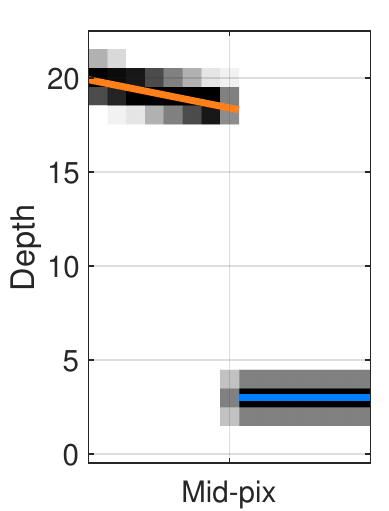} \vspace{-1mm}  \\
    \end{tabular}
    \end{center}
    \vspace{-2mm}
    \caption{\small An example of depth mixing, and how DC avoids it. (a) A slice through a depth image showing a depth discontinuity between two objects.  (b) An example sparse depth representation: each pixel either has a depth value or a zero. (c) The result of a 1D convolution, shown in (d), applied to the sparse depth.  This estimates the missing pixel, but generates a mixed-depth pixel between the two objects.  (e) A DC representation of the sparse depth.  Each pixel with a depth measurement has three non-negative coefficients that sum to $1$ (shown column-wise).  (f) The result of applying the same filter (d) to DC in (e).  Missing depths are interpolated and notably there is no depth mixing between the objects.}
    \label{fig:motiv_dc}\figvspace
\end{figure}

Depth completion aims to estimate unknown depths at image pixels using surrounding depth pixels plus the color image.  This is particularly challenging at depth discontinuities; here a foreground object occludes the background and the unknown pixel depths typically belong to either the foreground or background (see Fig.~\ref{fig:motiv_dc}).  
In this section we consider three contributing factors to depth mixing: depth ambiguity, the loss function, and depth representation.

\subsection{Depth Ambiguity}
\label{sec:ambiguity}

Depth completion often faces an ambiguity problem for pixels at object boundaries: do these pixels belong to the foreground or background object.  
To address this ambiguity, we first define the learning task: find model parameters $\theta$ that best predict depth $d_i$, given sparse depth and color image data $x_i$, with distribution $p_{data}$:
\begin{equation}
    \hat{\theta}=\argmax_\theta \mathbb{E}_{p_{data}}\left[ \log p_{model}(d_i|x_i;\theta) \right].
    \label{eq:optim}
\end{equation}
Here the expectation is performed over training data with distribution $p_{data}$.  The term $p_{model}$ is a probabilistic model for how well depth $d_i$ is predicted by data $x_i$.

Initially let's consider that the image data $x_i$ consist only of sparse depth values and no color images.  Given a sparsely sampled object, we can ask whether a pixel near a depth discontinuity boundary belongs to the foreground or background.  If the boundary is unknown,  it will be ambiguous whether it is foreground or background.  What this means in terms of Eq.~\ref{eq:optim} is that given the same data $x_i$, there are at least two compatible depths: $d^{(1)}$ for foreground and $d^{(2)}$ for background.  
If a color image is available, it may be possible to exactly infer the boundary between objects, and resolve this ambiguity.  However, often this is not the case; the boundary is not clear and hence the ambiguity persists.  In this paper we show that how this ambiguity is addressed has important implications for depth estimation.

\subsection{Depth Loss Functions with Ambiguity}

\begin{figure}[t!]
    \begin{center}
    \begin{tabular}{@{\hskip1pt}c@{\hskip1pt}c@{\hskip1pt}c@{\hskip1pt}c}
    ($a$) & \includegraphics[trim=30 0 50 38,clip,width=0.45\linewidth]{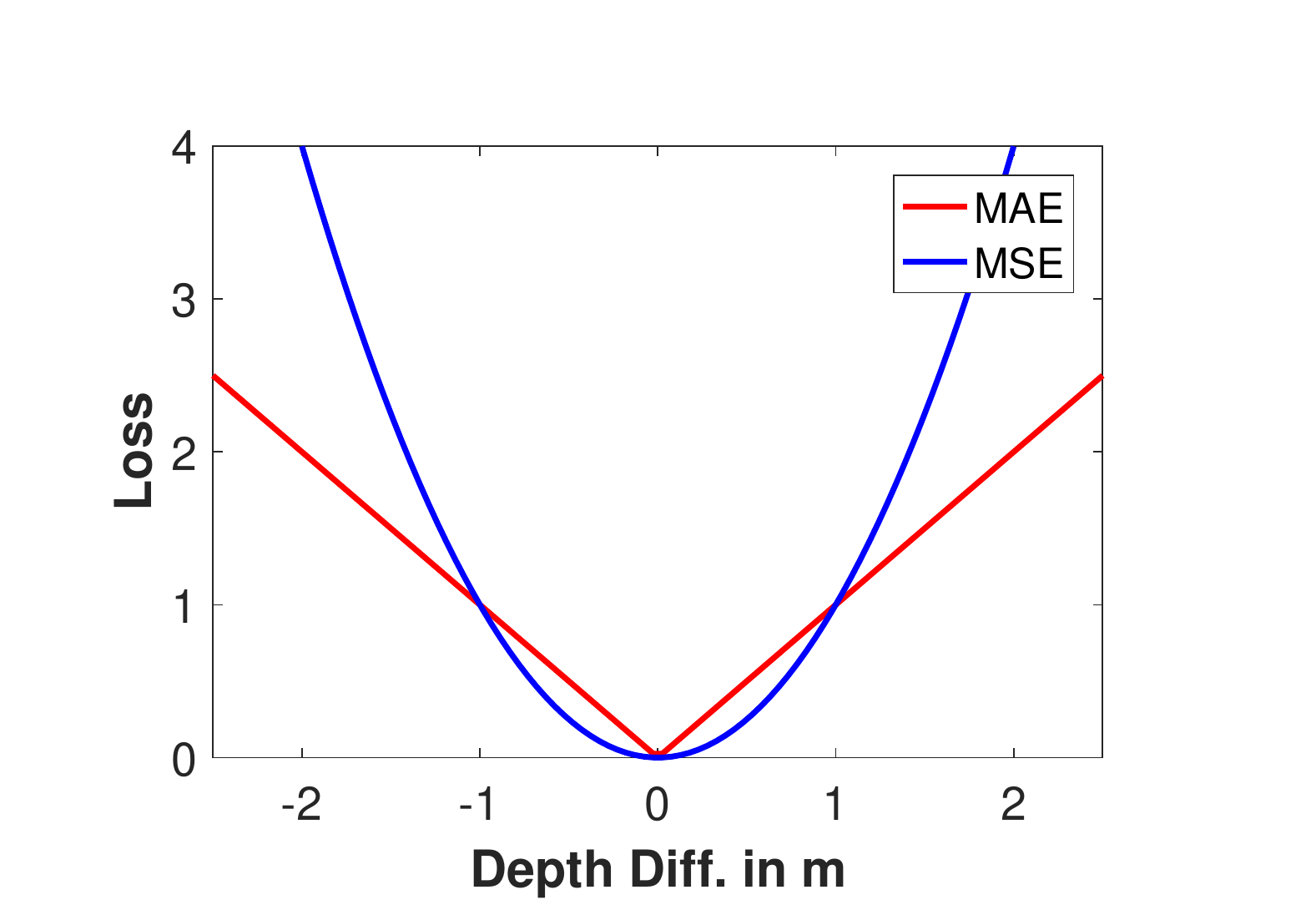} &
    ($b$) & \includegraphics[trim=30 0 50 38,clip,width=0.45\linewidth]{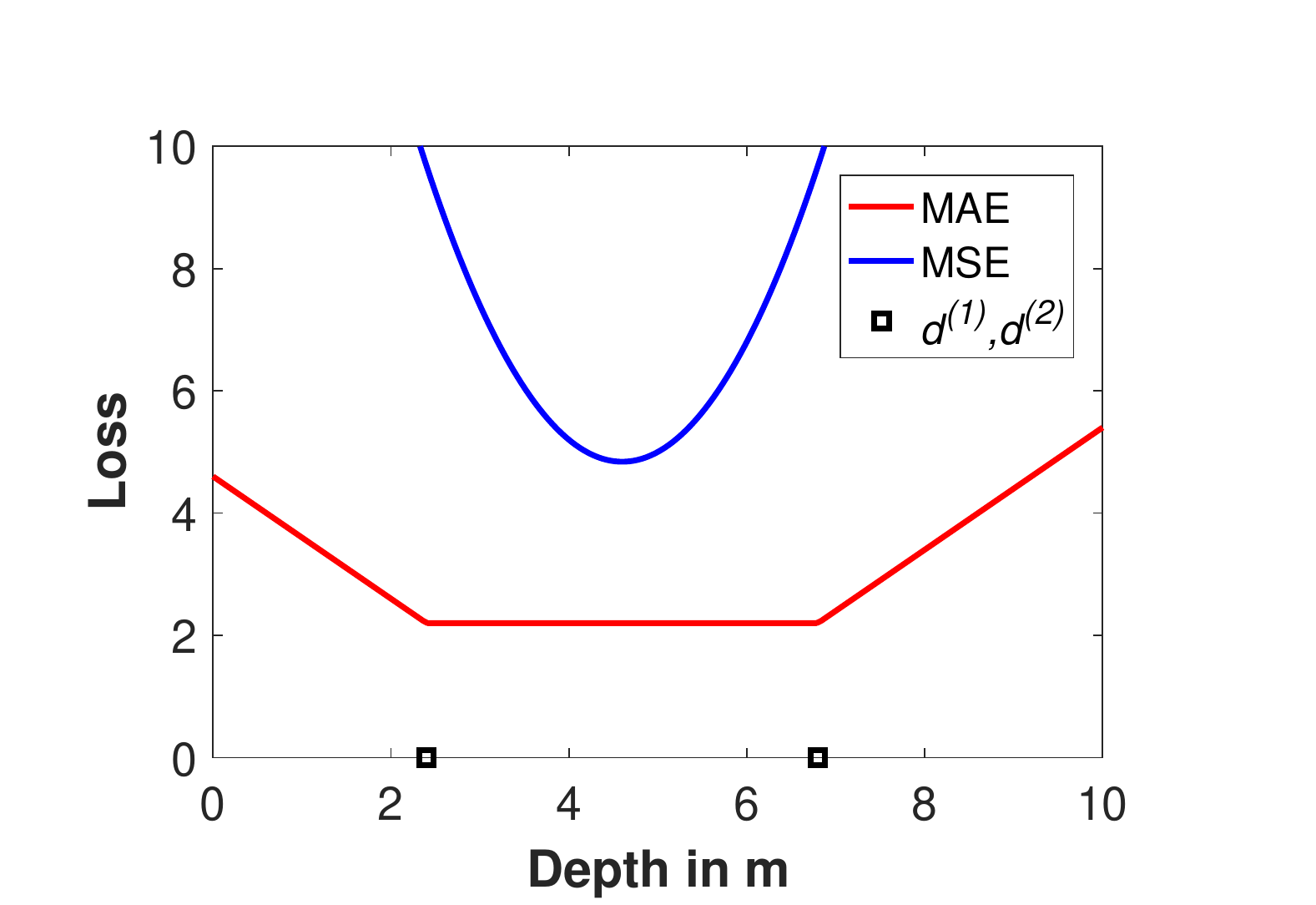} \\
    \end{tabular}
    \end{center}  \vspace{-2mm}
    \caption{(a) shows MSE and MAE loss functions. These perform an expectation over the probability of the data.  Now consider an ambiguous case where a pixel's depth has equal probability being $d^{(1)}$ or $d^{(2)}$, shown as black squares in (b). Minimum MSE estimate, $\hat{d}$, is the mid-point, while MAE has equal loss for all points between these two depths.  This illustrates why MSE prefers mixed-depth pixels, and MAE fails to penalize them.}
    \label{fig:mse_mae}\figvspace
\end{figure}

One of the more popular loss functions is Mean Squared Error (MSE). In part this is because the MSE gives the maximum likelihood solution to Eq.~\ref{eq:optim} when $p_{model}(d_i|x_i;\theta)$ is Gaussian.  Here we consider the implications of using MSE when there are depth ambiguities.  
First we address explicit ambiguities: there are two examples of data $x_i$, one having depth $d^{(1)}$ and the other $d^{(2)}$.  The MSE loss is:
\begin{equation}
    \mathrm{MSE}(d) = \frac{1}{2}\left(||d-d^{(1)}||^2 + ||d-d^{(2)}||^2 \right),
\end{equation}
which is minimum when
\begin{equation}
    \hat{d} = \frac{1}{2}(d^{(1)}+d^{(2)}).
\end{equation}
And so the estimated point is a mixed-depth pixel falling half-way between the foreground and background objects.  An illustration of this is in Fig.~\ref{fig:mse_mae}(b).

The same issue can occur even without explicit ambiguities.  
Assume we have a perfectly trained model that gives minimum MSE, and there are ambiguous situations as defined in Sec.~\ref{sec:ambiguity}.  Then as shown above, the minimum MSE solution will be for the model to predict mixed-depth pixels.  

Mean Absolute Error (MAE), has a similar issue, yet not as severe.  As in Fig.~\ref{fig:mse_mae}, in the pairwise ambiguity case, the MAE loss of mixed-depth pixels is equal to the loss at the actual values.  
Thus while MAE loss does not prefer mixed-depth pixels like MSE, nevertheless mixed-depth solutions may not be sufficiently penalized to avoid them.  

\subsection{Depth Evaluation}

RMSE~\footnote{RMSE has the same parametric minimum as MSE.} and MAE are used to evaluate depth completion.  This is concerning if there are depth ambiguities, since RMSE favors solutions with mixed-depth pixels.  
MAE, while not favoring these, may only penalize them weakly.  
Thus we need alternative evaluation metrics that properly penalize mixed-depth pixels.

\begin{figure}[t!]
    \centering
    \includegraphics[trim=120 10 220 0,clip, width=\linewidth]{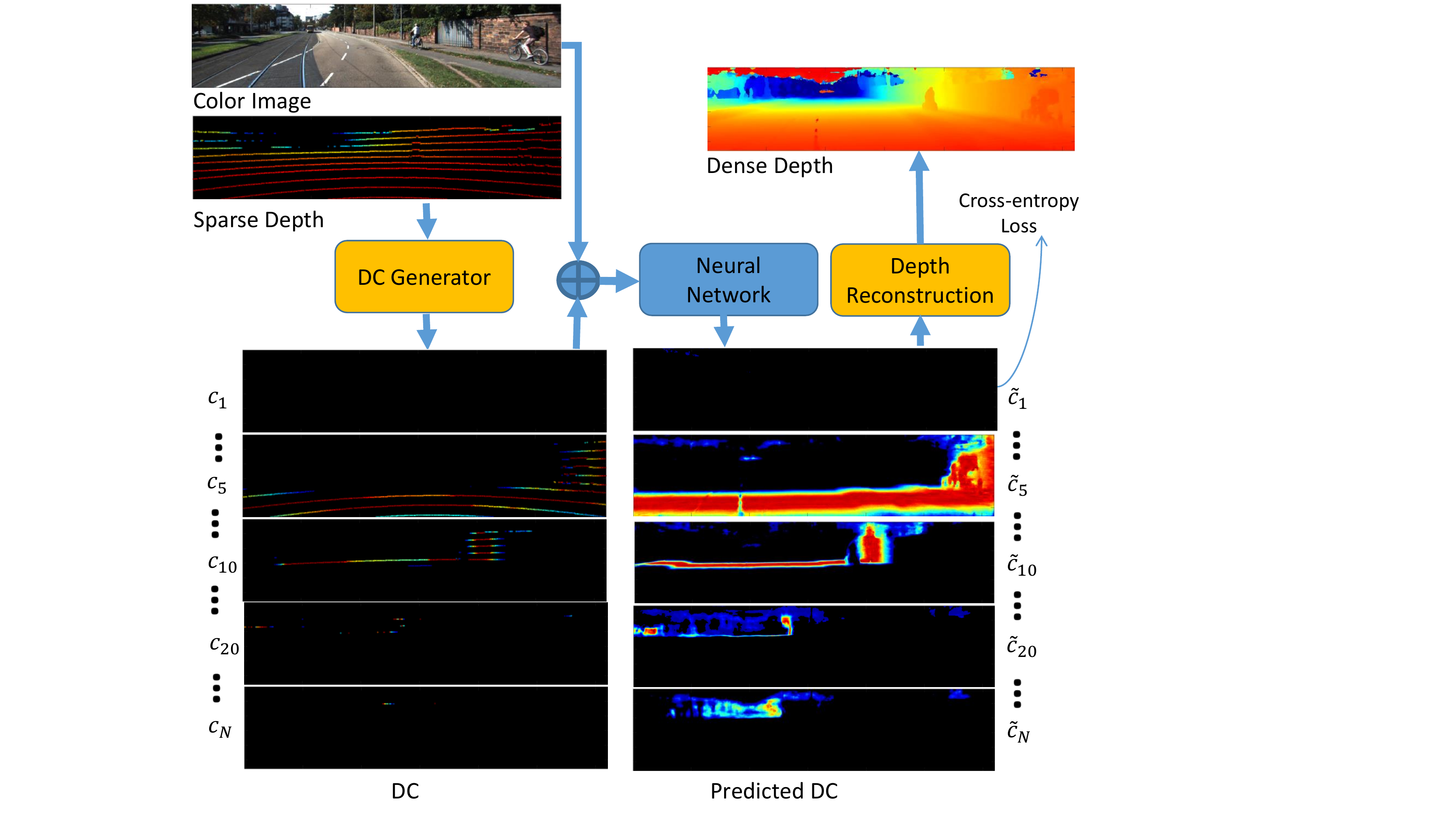}
    \caption{An overview of our method. Sparse depth is converted into Depth Coefficients with multiple channels, each channel holding information of a certain depth range.  This, along with color, is input to the neural network.  The output is a multi-channel dense depth density that is optimized using cross entropy with a ground-truth DC. The final depth is reconstructed based on the predicted density.}
    \label{fig:overview}\figvspace
\end{figure}

\subsection{Depth Representation}

An important factor impacting depth mixing, is how depth is represented when input into a depth completion CNN.  The usual representation, which we call a sparse depth image, is to project known depth points into the image plane, setting the pixel value at that location to the depth, and setting the remaining pixels to zero. 
The CNN applies a sequence of convolutions and non-linear operations to the sparse depth image eventually predicting a dense depth image.  To understand the tendency of CNNs to generate mixed depths, consider a slice through a simple two-object scene illustrated in Fig.~\ref{fig:motiv_dc}(a-d). 
Applying a smoothing convolution from (d) generates the dense depth estimate in (c). 
Notice the predicted center pixel depth is an intermediate of the two object depths -- an example of depth mixing creating a spurious point in empty space.

To avoid depth mixing, the mid-pixel in Fig.~\ref{fig:motiv_dc}(c) should be predicted purely from the (foreground or background) object to which it belongs.  In $1$D it is simple to find convolutions that avoid depth mixing, e.g., averaging to the right or left.  But for $2$D depth images with unevenly distributed measurements, it is much more complicated to avoid mixing, as depth boundaries can have many shapes with respect to known depth pixels.  At the very least, doing so requires learning a complex network.  
The simple alternative, presented next, is to use a representation where convolutions can directly generate hypotheses without depth mixing.


\section{Methodology}

This section provides a two-part approach to address the depth mixing problem in depth completion.  
The first part is to use an alternative representation for depth.  
This representation enables use of cross-entropy loss, which overcomes the issues described in Sec.~\ref{sec:mixing}.
What we propose is not a new CNN architecture, rather it is a modification to the input and output of a broad class of architectures for depth completion. These modifications aim to improve the performance of otherwise good algorithms by reducing pixel depth mixing. Fig.~\ref{fig:overview} shows the overview of our approach.

\subsection{Depth Coefficients}

We seek a depth representation that not only can easily avoid pixel depth mixing, but also enables interpolation between depths within objects.  
One solution is to use a discrete one-hot depth representation.  
Despite enabling convolutions without depth mixing, its drawback is the trade-off between a loss in depth accuracy and a very large number of channels to represent depth. 
Here we propose an alternative that has the benefits of one-hot encoding, but requires far fewer channels and eliminates the accuracy loss issue.  

To represent a dense or sparse depth image we create a multi-channel image of the same size, with each channel representing a fixed depth, $D=\{D_1,\ldots,D_N\}$. 
The depth values increase in even steps of size $b$.  
In choosing the number of channels (or bins) we trade-off memory vs.~precision.
For our applications, we chose $80$ bins to span the full depth being modeled, and this determines the bin width, $b$. 
Thus each pixel $i$ has a vector of values, $\bm{c}_i=\{c_{i1},\ldots,c_{iN}\}$, which we call {\it Depth Coefficients} (DC), that represents its depth, $d_i$.  
We constrain these coefficients to be non-negative, sum to $1$, and give the depth as the inner product with the channel depths:
\eqnvspace
\begin{equation}
    d_i = \sum_j c_{ij}D_j.
    \label{eq:DCP}\eqnvspace
\end{equation}
Note this representation is not unique as many combinations of coefficients may produce the same depth. 

So we use the following simple, sparse representation with three non-zero coefficients to represent depth. Let $k$ be the index of the depth channel closest to pixel depth $d_i$ and $\delta=\frac{d_i-D_k}{b}$. Since $b$ is the spacing between adjacent bin depths, the $D_{k-1}$ and $D_{k+1}$ bins can be expressed in terms of the center bin as $D_{k-1}=D_k-b$ and $D_{k+1}=D_k+b$ respectively. With this substitution all the terms cancel on the right-hand side of Eq.~\ref{eq:dcCalc} leaving $d_i$. The DC vector for pixel $i$ is:
\eqnvspace
\begin{equation}
    \bm{c}_{i} = (0,\ldots,0,\frac{0.5-\delta}{2}, 0.5, \frac{0.5+\delta}{2},0,\ldots,0),
    \label{eq:dcCalc}
\end{equation}
where three non-zero terms are $(c_{i(k-1)},c_{ik},c_{i(k+1)})$.  This is unique for each $d_i$, satisfies Eq.~\ref{eq:DCP}, and sums to 1.

More than representing a continuous value as weighted sum of discrete bins \cite{zeisl2014discriminatively}, we claim that
using DC to represent depth provides a much simpler way for CNNs to avoid depth mixing. 
The first step of a CNN is typically an image convolution with $N_{in}$ input channels. 
For sparse depth input, $N_{in}=1$, and so all convolutions apply equally to all depths, resulting mixing right from the start.
For DC input, depths are divided over $N_{in}=N$ input channels, resulting in two important capabilities.  First, CNNs can learn to avoid mixing depths in different channels as needed.  This is similar to voxel-based convolutions \cite{ghadai2018multi,maturana2015voxnet} which avoid mixing spatially-distant voxels.  This effect is illustrated in Fig.~\ref{fig:motiv_dc}(e-f), where a multi-channel input representation, (e), allows convolutions to avoid mixing widely spaced depths.  Second, since convolutions apply to all channels simultaneously, depth dependencies, like occlusion effects, can be modeled and learned by neural networks.

\subsection{Loss Function}
\label{sec:crossEntropy}


As shown in Sec.~\ref{sec:mixing}, MSE leads to depth mixing when there is depth ambiguity.
One way to avoid this is, rather than estimate depth directly, to estimate a more general probabilistic representation of depth. 
Now DC can provide a probabilistic depth model, both for $p_{data}$ and $p_{model}$ in Eq.~\ref{eq:optim}.  Minimizing the cross entropy of the predicted output $\tilde{\bm{c}}$, representing $p_{data}(\tilde{d_i}|x_i;\theta)$, is equivalent to minimizing the KL divergence with $\bm{c}$. In this way, we can learn to estimate $p_{model}(d_i|x_i;\theta)$ parameterized with DC.  
Our cross-entropy loss for pixel $i$ is defined as:
\eqnvspace
\begin{equation}
    L_i^{ce}(c_{ij}) = -\sum_{j=1}^N c_{ij}\log {\tilde{c}_{ij}},
    \label{eq:crossEntropy}
\end{equation}
where $c_{ij}$ terms are the DC elements of the ground truth obtained using Eq.~\ref{eq:dcCalc}.  
Training a network to predict $\tilde{c}_{ij}$ that minimizes $L_i^{ce}$ is equivalent to maximizing Eq.~\ref{eq:optim}.  

\begin{figure}
    \centering
    \includegraphics[trim=10 0 10 0,clip,width=0.9\linewidth]{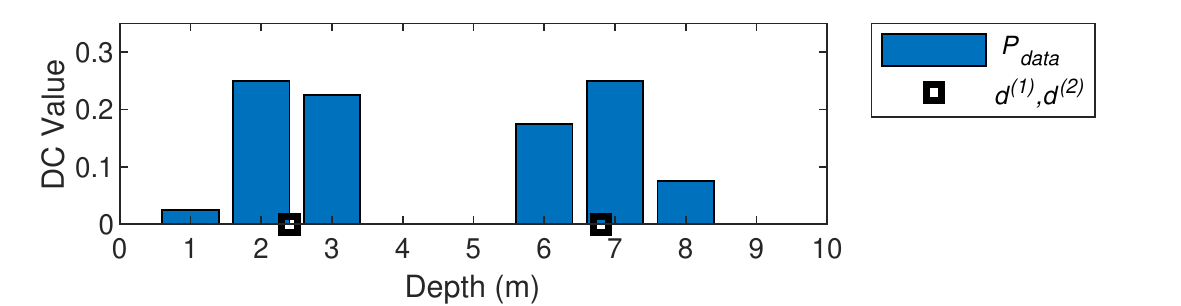}
    \caption{An illustration of $P_{data}$ modeled as the sum of the DC of the two points from Fig.~\ref{fig:mse_mae}.  The estimated $\hat{c}_{ij}$ with minimum cross-entropy loss, Eq.~\ref{eq:crossEntropy}, will exactly match $P_{data}$, providing a multi-modal density.  A pixel depth estimate using Eq.~\ref{eq:estDC} will find the depth of one of the peaks, and not a mixed-depth value.  }
    \label{fig:crossEntropy} \figvspace
\end{figure}

Use of cross-entropy loss has two main advantages.  The first is that depth ambiguities no longer result in a preference for mixed-depth pixels.  As illustrated in Fig.~\ref{fig:crossEntropy}, DC models multi-modal densities, and as we show in the next section our depth estimate will find the location of the maximum peak at one of the depths.  Second, optimizing cross entropy leads to much faster convergence than MSE, which suffers from gradients going to zero near the solution.

\subsection{Depth Reconstruction}
\begin{figure}[ht!]
    \centering
    \begin{tabular}{@{\hskip1pt}c@{\hskip1pt}c@{\hskip1pt}c@{\hskip1pt}c}
    ($a$) &
    \includegraphics[trim=25 0 50 38,clip,width=0.45\linewidth]{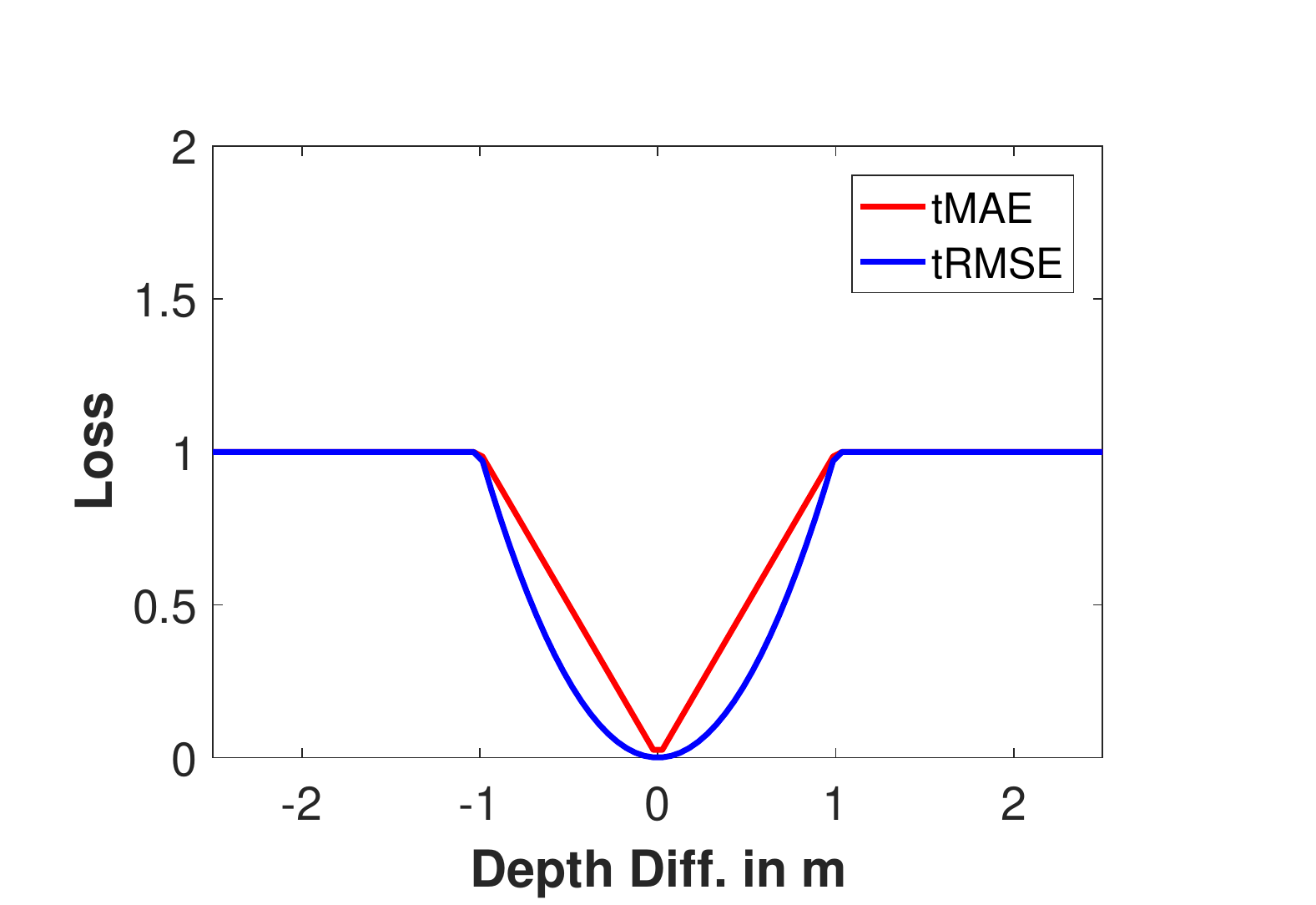} & ($b$)
    \includegraphics[trim=25 0 50 38,clip,width=0.45\linewidth]{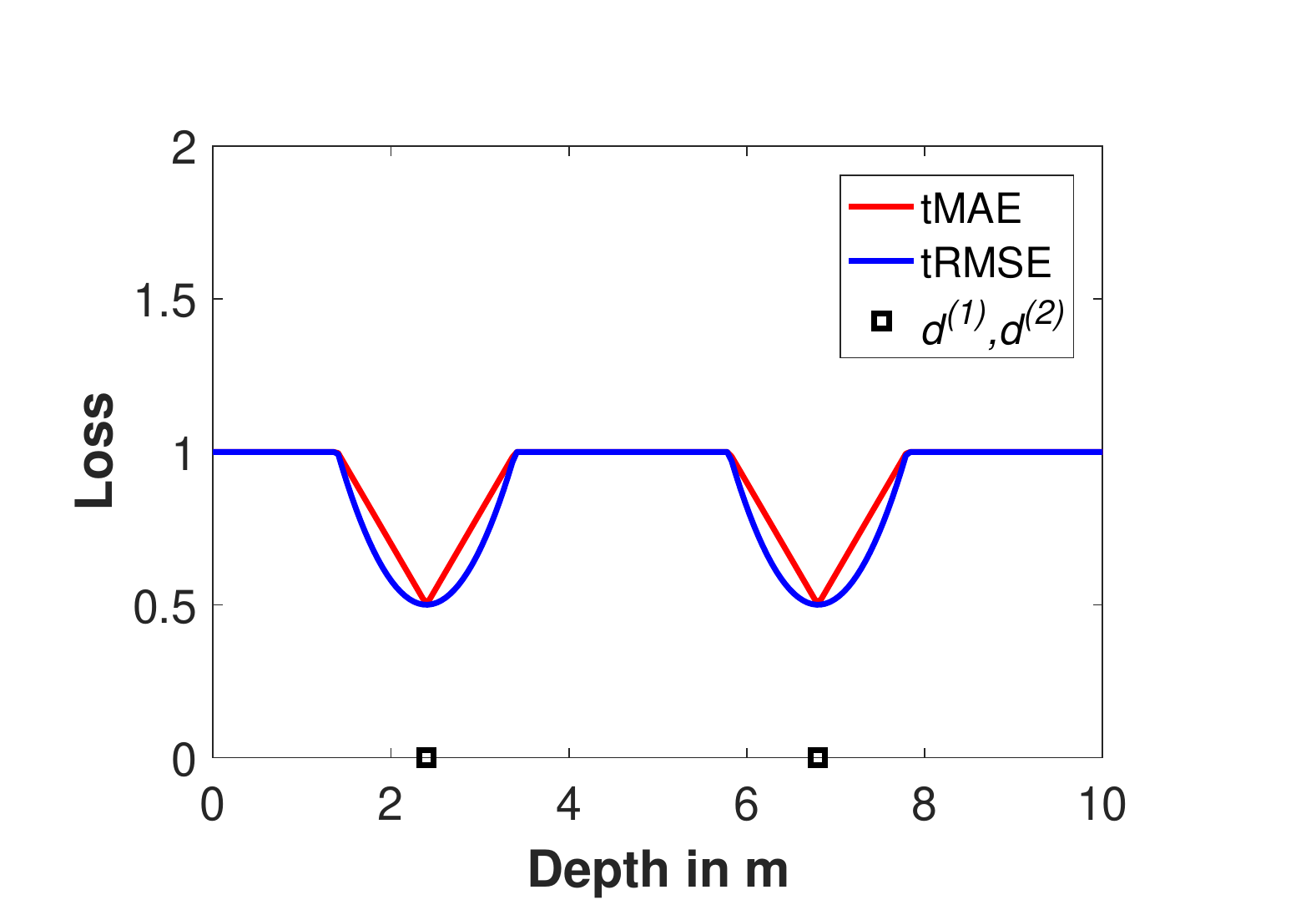} 
    \end{tabular}
    \caption{(a) tRMSE and tMAE with threshold $t=1$, compare to MSE and MAE in Fig.~\ref{fig:mse_mae}(a).  (b) When there is a depth ambiguity of at least $t$, in each case the minima will be at the true depth and the ambiguous depth, while the mixed-depth region between them will be penalized equally to other large-error regions. }
    \label{fig:rtmse}\figvspace
\end{figure}
There are a number of options for depth reconstruction.  We can use Eq.~\ref{eq:DCP}, and substitute $\hat{c}_{ij}$ for $c_{ij}$ for pixel $i$.  However, the predicted coefficients may be multi-modal as in Fig.~\ref{fig:crossEntropy}, and it may be preferable to estimate the maximum likelihood solution.  We can estimate the depth for the peak via the maximum coefficient $c_{ik}\in\bm{c}_i$ and its two neighbors:
\eqnvspace
\begin{equation}
    \hat{d}_i = \frac{\hat{c}_{i(k-1)} D_{(k-1)} + \hat{c}_{ik} D_{k} + \hat{c}_{i(k+1)} D_{(k+1)} }{\hat{c}_{i(k-1)} + \hat{c}_{ik} + \hat{c}_{i(k+1)} }.
    \label{eq:estDC}
\end{equation}
A third way is for the network to directly predict depth in addition to DC.

\subsection{New Evaluation Metrics}
\begin{figure*}[hbtp!]
    \centering
    \includegraphics[trim= 2 300 0 60,clip,width=0.95\linewidth]{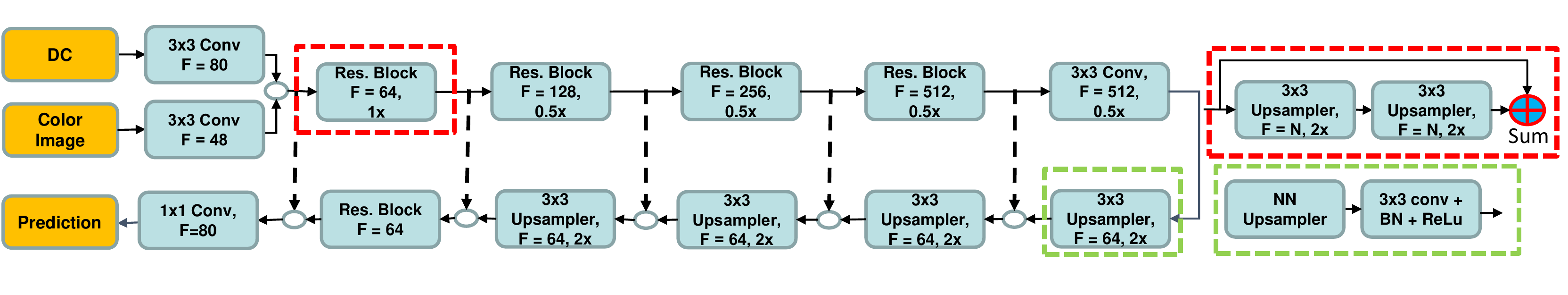}
    \caption{Our CNN architecture modified from~\cite{ma2018self} with $80$-channel DC at the input, and $80$-channel cross-entropy loss. The residual blocks are defined in the dashed red region, and the top branch consists of ResNet $34$~\cite{He:2016:DeepResNet}. }
    \label{fig:nn_architecture}
    \vspace{-0.3pt}
\end{figure*}

\begin{table*}[hbtp!]
    \centering
    \setlength\doublerulesep{0.5pt}
    \begin{tabular}{|c|c|c|c|c|c|c|c|c|c|c|}
    \hline
        Method & RMSE & MAE & REL & tMAE & tRMSE & $\delta_1$ & $\delta_2$ & $\delta_3$ & $\delta_4$ & $\delta_5$ \\
        \hhline{===========}
        Ma \cite{mal2018sparse} & $0.236$ & $0.13$ & $0.046$ & $0.068$ & $0.075$ & $52.3$ & $82.3$ & $92.6$ & $97.1$ & $99.4$ \\ \hline
        Bilateral \cite{BarronPoole2016} & $0.479$ & - & $0.084$ & - & - & $29.9$ & $58.0$ & $77.3$ & $92.4$ & $97.6$ \\ \hline
        SPN \cite{liu2016learning} & $0.172$ & - & $0.031$ & - & - & $61.1$ & $84.9$ & $93.5$ & $98.3$ & $99.7$ \\ \hline
        Unet \cite{cheng2018depth} & $0.137$ & $0.051$ & $0.020$ & - & - & $78.1$ & $91.6$ & $96.2$ & $98.9$ & $99.8$\\ \hline
        CSPN \cite{cheng2018depth} & $0.162$ & - & $0.028$ & - & - & $64.6$ & $87.7$ & $94.9$ & $98.6$ & $99.7$\\ \hline
        CSPN+UNet \cite{cheng2018depth} & \boldsymbol{$0.117$} & - & $0.016$ & - & - & $83.2$ & $93.4$ & $97.1$ & $99.2$ & \boldsymbol{$99.9$} \\
        \hline
        Ours-all & $0.118$ & \boldsymbol{$0.038$} & \boldsymbol{$0.013$} & 0.042 & \boldsymbol{$0.053$} & $86.3$ & $95.0$ & $97.8$ & \boldsymbol{$99.4$} & \boldsymbol{$99.9$} \\ \hline
        Ours-3coeff & $0.131$ & \boldsymbol{$0.038$} & \boldsymbol{$0.013$} & \boldsymbol{$0.040$} & $0.054$ & \boldsymbol{$86.8$} & \boldsymbol{$95.4$} & \boldsymbol{$97.9$} & $99.3$ & $99.8$ \\\hline
          
    \end{tabular}
    \caption{Quantitative results of NYU2 (Done on Uniform-$500$ Samples + RGB) (units in m).}
    \label{tab:nyu2_results}
\end{table*}
While RMSE and MAE are useful metrics for overall depth completion performance, we showed in Sec.~\ref{sec:mixing} that these encourage, or at least do not sufficiently penalize, depth mixing.  
Thus we propose two complementary metrics that focus on depth surface accuracy and penalize depth mixing equally to other large errors.  
These metrics are Root Mean Squared Thresholded Error (tRMSE) and Mean Absolute Thresholded Error (tMAE), defined as follows:
\eqnvspace
\begin{equation}
    \mathrm{tRMSE} = \sqrt{\sum_{i=1}^P \frac{\mathrm{min}( (\tilde{y}_i-\hat{y}_i)^2, t^2 )}{P}},
\end{equation}
\eqnvspace
\begin{equation}
    \mathrm{tMAE} = \sum_{i=1}^P  \frac{\mathrm{min}( |\tilde{y}_i-\hat{y}_i|, t )}{P}.
\end{equation}
Here $P$ is the number of pixels, $t$  the threshold distance distinguishing within-surface variation from inter-object separation, $\tilde{y}_i$ the ground-truth value and $\hat{y}_i$ the estimated value.

To observe the advantage of tRMSE and tMAE, consider a single pixel with a density function defined over its depth.  Assume that this density is greater than zero for a set of points (corresponding to probable surfaces) and zero elsewhere.  Now all depths, separated by at least $t$ from any probable depth, will have a fixed cost of $t$ which is greater than the cost of depths closer than $t$ to a probable depth.  Unlike RMSE, there are no local minima at mixed-depth points ($>t$ from a probable point) as described in Sec.~\ref{sec:mixing}. Unlike MAE, all mixed-depth points have greater cost than points close to probable surfaces. We illustrate this property in Fig.~\ref{fig:rtmse}. While we propose tRMSE and tMAE as new metrics, we do not advocate them as loss functions. This is because at large errors the gradients are zero, and we observe very slow convergence.  Rather, cross-entropy on DC is a suitable loss function as described in Sec.~\ref{sec:crossEntropy}.

\subsection{Uses of Depth Completion}

Ultimately the choice of algorithm and evaluation metric should depend on the use of depth completion. We identify two uses where depth mixing can have a significant impact.  The first is to create dense, pixel-colored, 3D environment models from lower-resolution depth sensors.  Now mixed-depth pixels occurring in empty space between objects, illustrated in Fig.~\ref{fig:mse_mae}, can create connecting surfaces and negatively impact the visual quality of the 3D environment.

Another use of depth completion is for data fusion and subsequent tasks such as object detection.  If effective, this could enable low-cost, low-resolution sensors to be upgraded when combined with a color camera.  We compare object detection performance with super-resolved depths both with and without our contribution. 

\vspace{-0.5cm} \paragraph{Architecture}  This work explores using DC for the input and output of a CNN.   Thus we select a standard network~\cite{ma2018self} and adjust the input and output blocks to use DC, and train with cross-entropy loss (see Fig.~\ref{fig:nn_architecture}).

\section{Experiments}
\subsection{Experimental Protocols}
We evaluate DC representation by means of two publicly available datasets: KITTI (outdoor scenes) and NYU2 (indoor scenes) respectively to demonstrate the performance of our algorithm. 
We use KITTI depth completion dataset~\cite{Uhrig2017THREEDV} for both training and testing. The dataset is created by aggregating Lidar scans from $11$ consecutive frames into one, producing a semi-dense ground truth with roughly $30\%$ annotated pixels. The dataset consists of $85,898$ training data, $1,000$ selected validation data, and $1,000$ test data without ground truth. We truncate the top $90$ rows of the image during training since it contains no Lidar measurements. 

The NYU-Depth v2 dataset consists of RGB and depth images collected from $464$ different scenes. We use the official split of data, where $249$ scenes are used for training and we sample $50$K images out of the training similar to~\cite{mal2018sparse}. For testing, the standard labelled set of $654$ images is used. The original image size is first downsampled to half, and then center-cropped, producing a network input spatial dimension of $304 \times208$.  For comparison purposes, we choose the state of the arts in both outdoor~\cite{ma2018self} and indoor scenes~\cite{mal2018sparse, cheng2018depth} using RGBD depth sensors.

\begin{table}[t!]
    \centering
    \setlength\doublerulesep{0.5pt}
    \resizebox{\linewidth}{!}{
    \begin{tabular}{|c|c|c|c|c|c|c|}
    \hline
        Method & MAE & RMSE & iMAE & iRMSE & tMAE & tRMSE \\
        \hhline{=======}
        Ma~\cite{ma2018self} & $65.2$ & $174.3$ & - & - & $59.4$ & $69.5$ \\ 
        \hline
        DC-all & $38.6$ & \boldsymbol{$142.3$} & $1.55$ & \boldsymbol{$2.23$} & $36.1$ & $50.9$ \\
        \hline
        DC-3coeff & \boldsymbol{$37.8$} & $160.6$ & \boldsymbol{$1.53$} & 2.41 & \boldsymbol{$33.4$} & \boldsymbol{$47.2$} \\
        \hline
    \end{tabular}}
    \caption{Depth completion results on KITTI validation benchmark with $16$-row Lidar input (units cm).}
    \label{tab:kitti_results_valid}
\end{table}

\vspace{-0.5cm} \paragraph{Sub-Sampling}
\begin{figure*}[ht!]
    \begin{center}
      \begin{tabular}{@{\hskip1pt}c@{\hskip1pt}c@{\hskip1pt}c@{\hskip1pt}c@{\hskip1pt}c@{\hskip1pt}c@{\hskip1pt}c@{\hskip1pt}c}
      ($a$) &
        \includegraphics[trim=300 230 250 100,clip,width=0.36\linewidth]{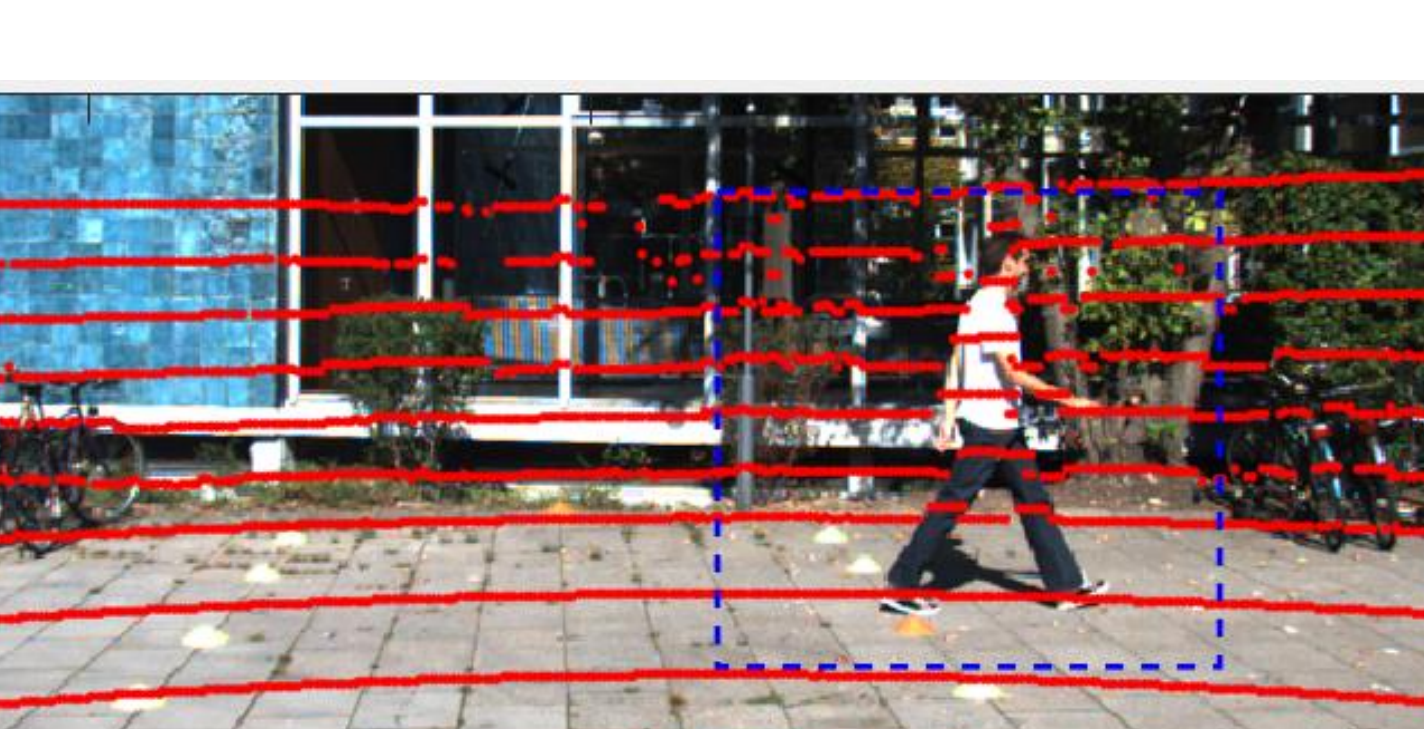} & ($b$) & 
        \includegraphics[trim=345 220 350 50,clip,width=0.16\linewidth]{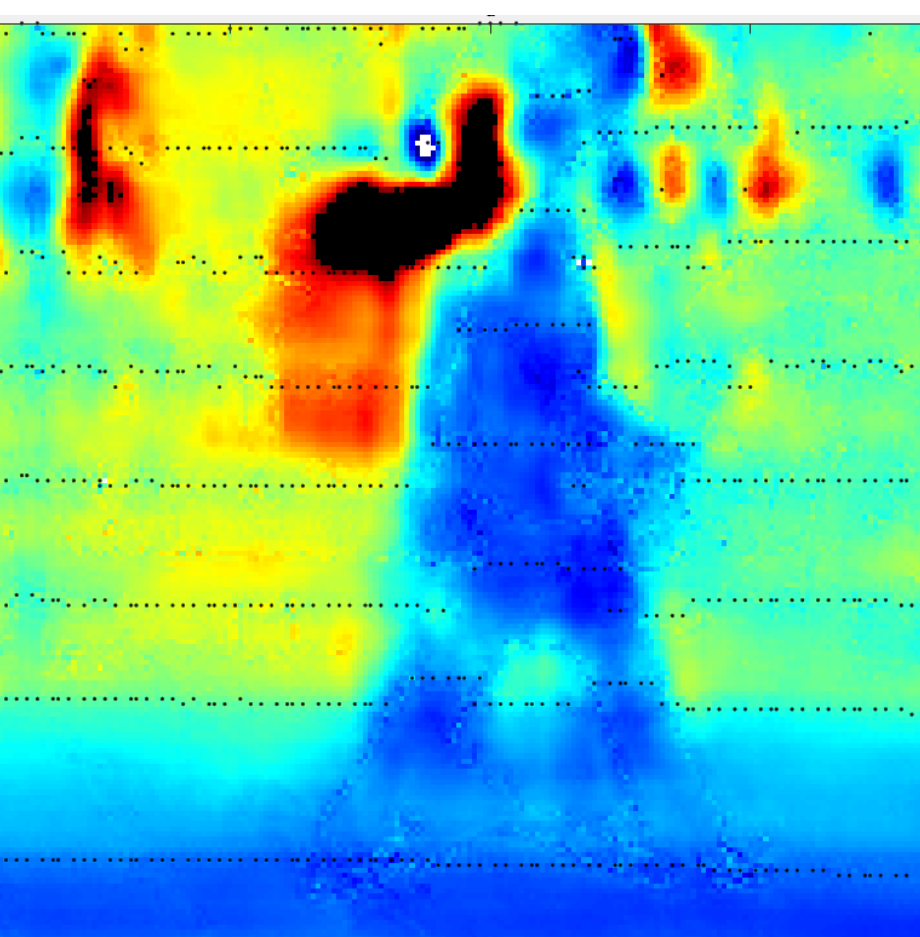} 
        & ($c$) & \includegraphics[trim=345 220 350 50,clip,width=0.16\linewidth]{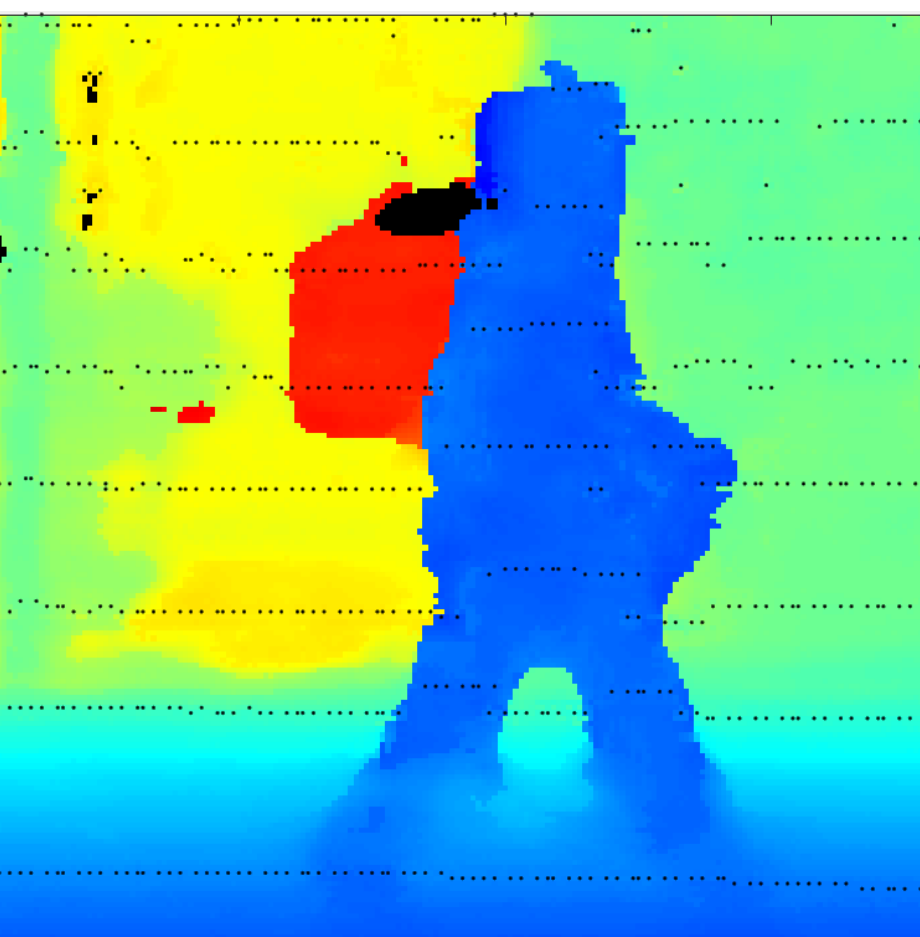} 
        & ($d$) & \includegraphics[trim=345 220 350 50,clip,width=0.16\linewidth]{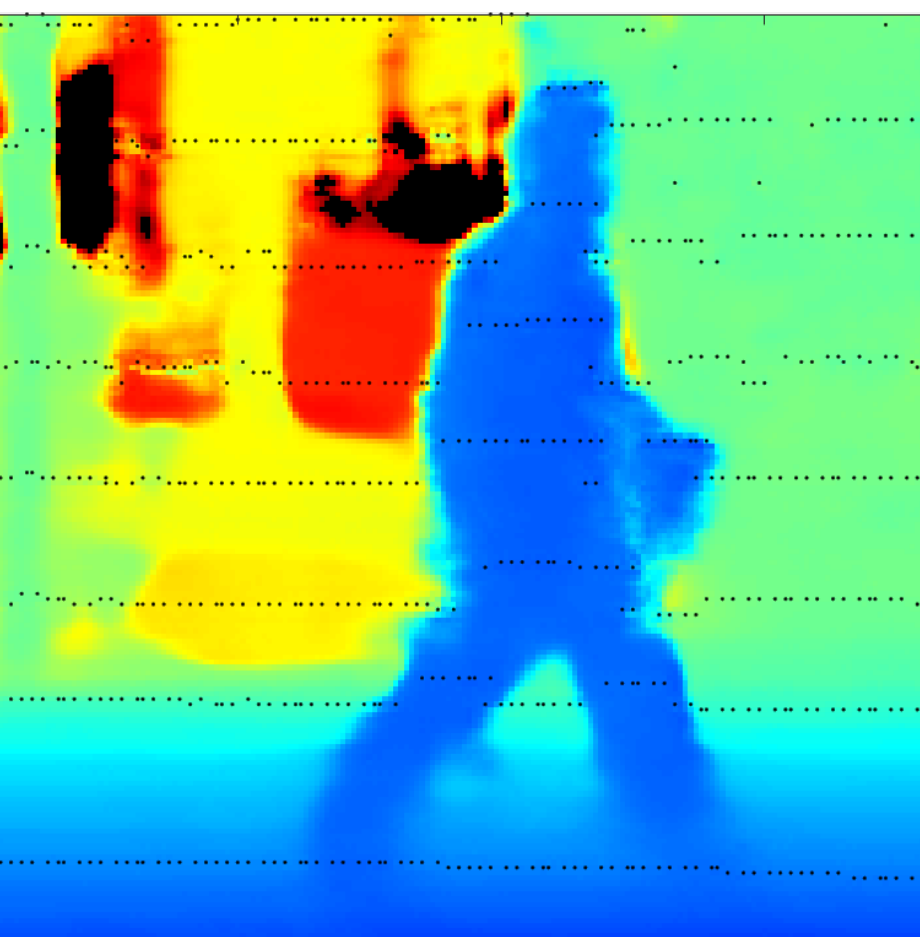} \\
       \end{tabular}
         \begin{tabular}{@{\hskip1pt}c@{\hskip1pt}c@{\hskip1pt}c@{\hskip1pt}c@{\hskip1pt}c@{\hskip1pt}c}
            ($e$) &\includegraphics[trim=10 40 20 70,clip,width=0.29\linewidth]{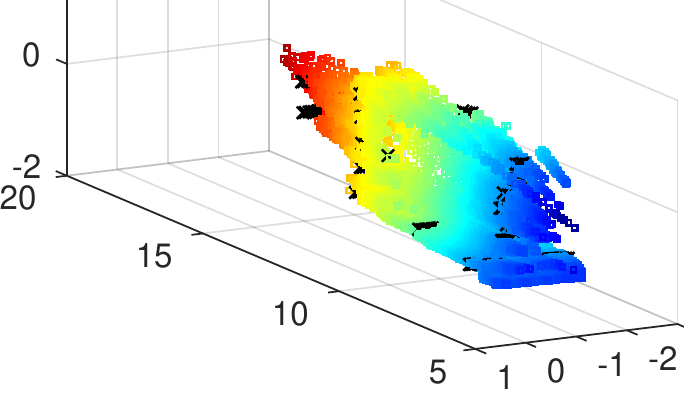} &  ($f$) &
            \includegraphics[trim=10 40 20 70,clip,width=0.29\linewidth]{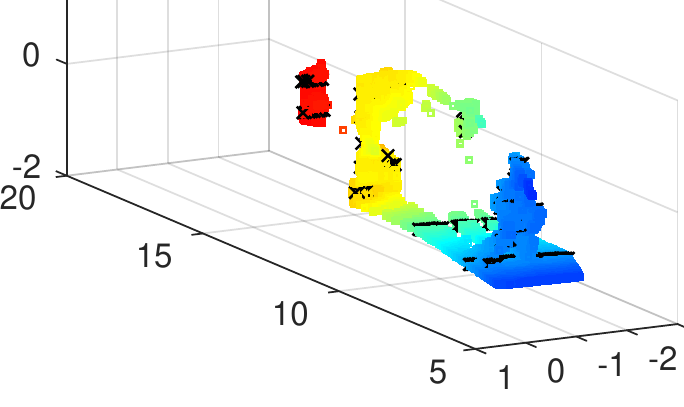} &  ($g$) &
            \includegraphics[trim=10 40 20 70,clip,width=0.29\linewidth]{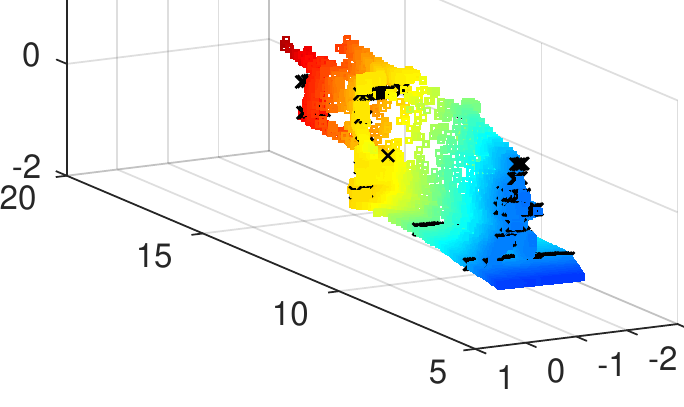}\\
              \end{tabular}
        \end{center}
          \vspace{-3mm}
    \caption{Depth completion with $16$-row Lidar.  ($a$) scene, ($b,e$) show Ma {\em et al.}~\cite{ma2018self} with significant mixed pixels.  ($c,f$) show our $3$-coefficient estimation, demonstrating very little depth mixing.  ($d,g$) show our estimation with all coefficients.}
    \label{fig:pedestrian}\figvspace
\end{figure*}

\begin{figure*}[ht!]
   \begin{center}
    \begin{tabular}{@{\hskip1pt}c@{\hskip1pt}c@{\hskip1pt}c@{\hskip1pt}c@{\hskip1pt}c@{\hskip1pt}c@{\hskip1pt}c@{\hskip1pt}c}
        ($a$) &  \includegraphics[trim=200 200 350 110,clip,width=0.22\linewidth]{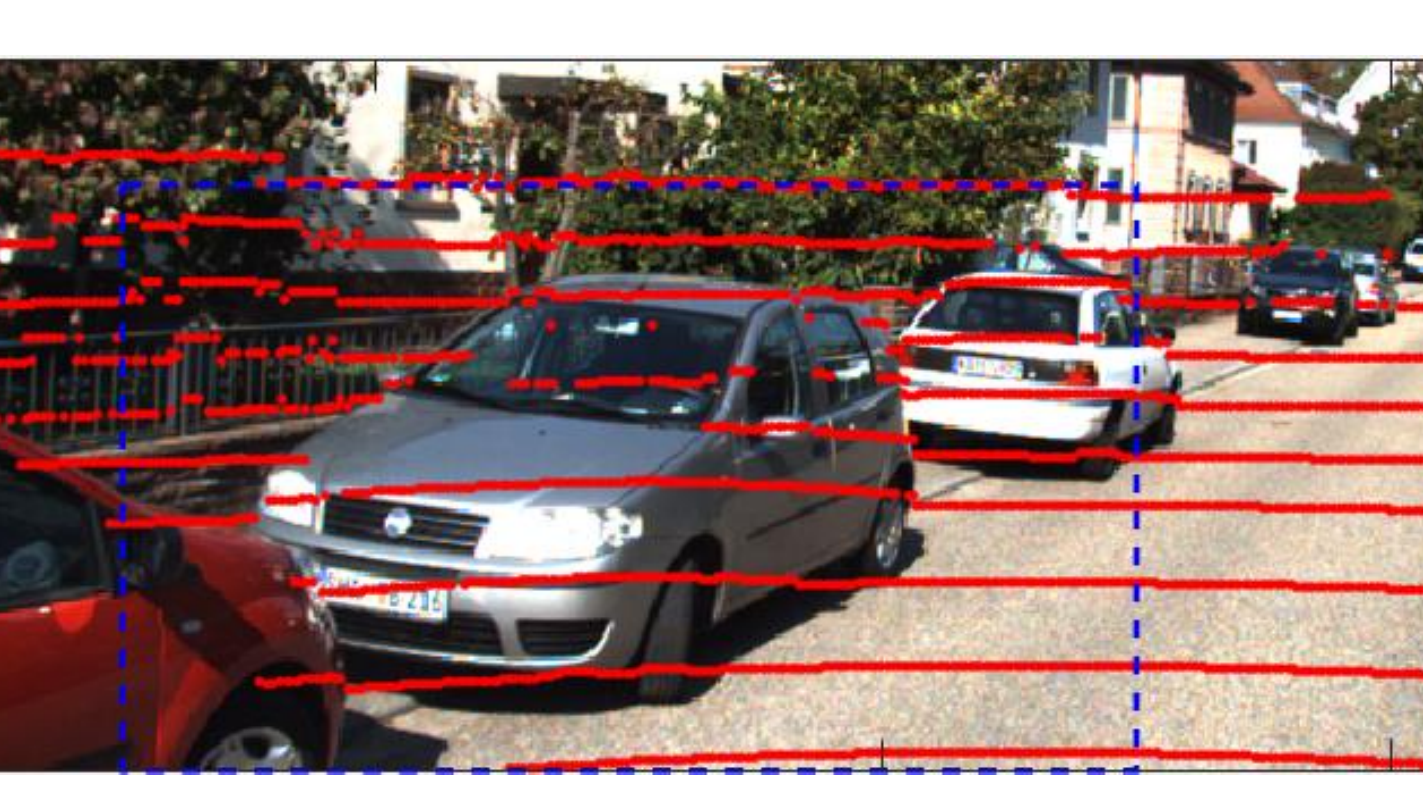} & ($b$) &
         \includegraphics[trim=260 200 300 90,clip,width=0.22\linewidth]{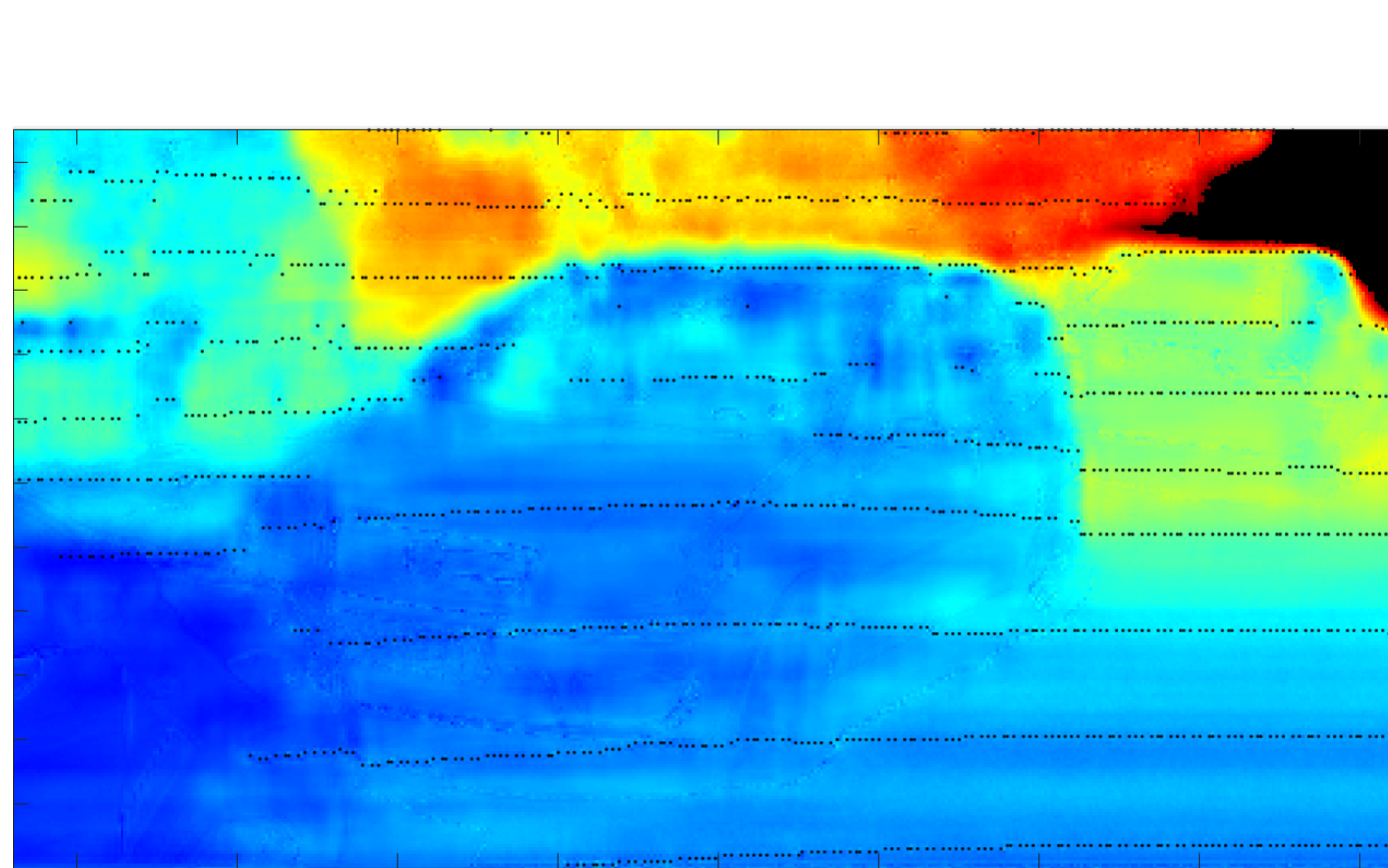} & ($c$) &
         \includegraphics[trim=260 200 300 90,clip,width=0.22\linewidth]{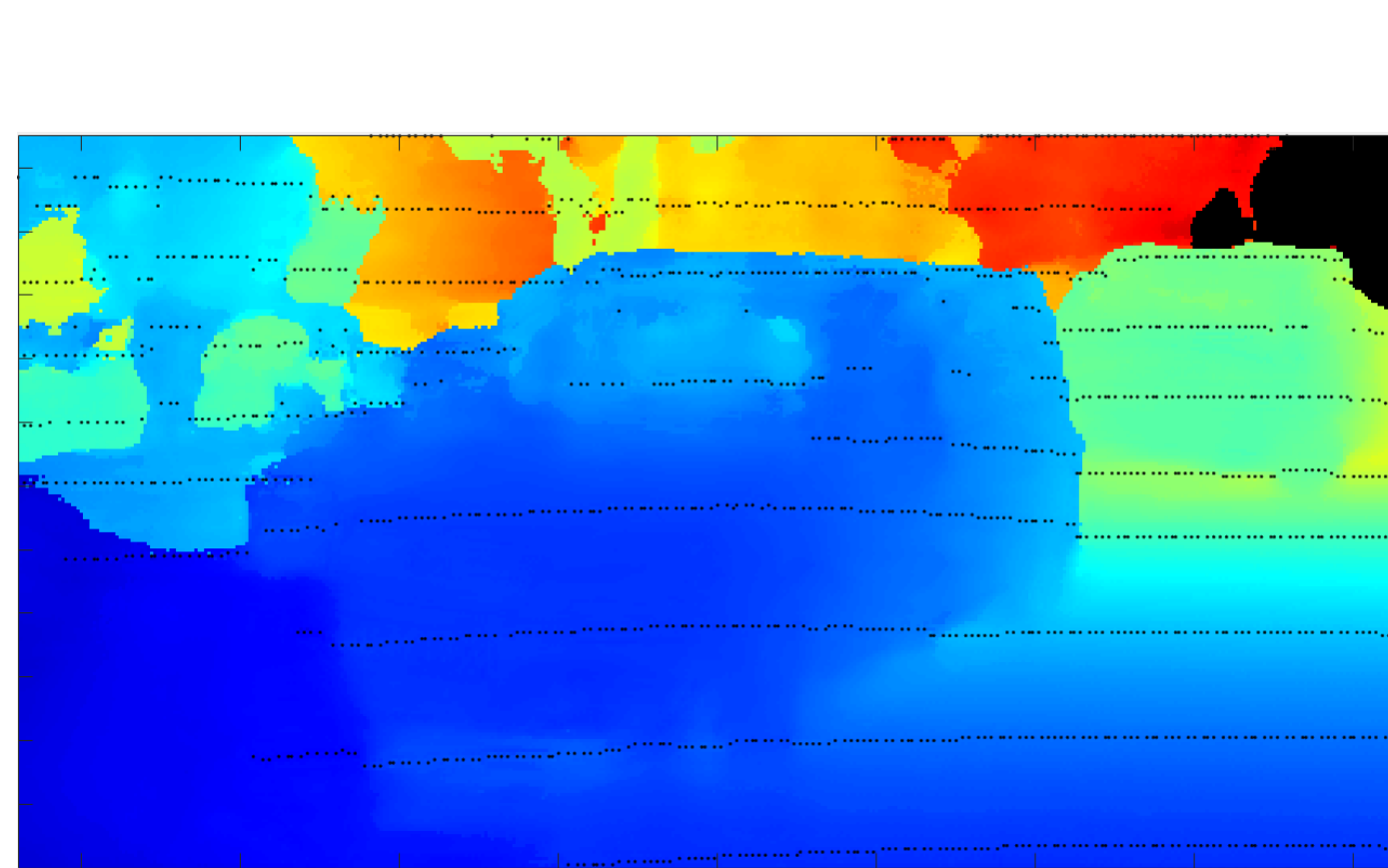} & ($d$) &
         \includegraphics[trim=260 200 300 90,clip,width=0.22\linewidth]{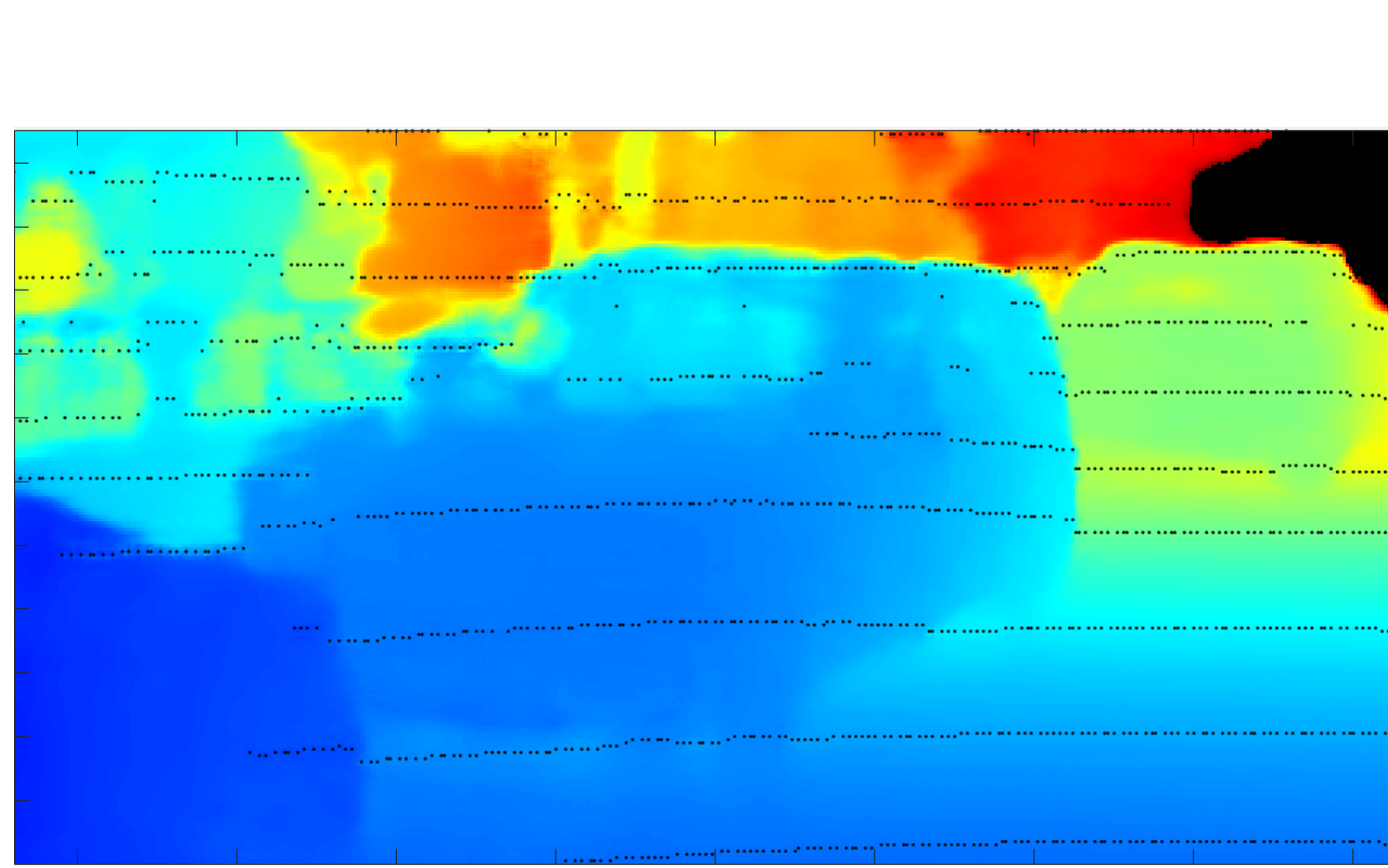}
          \end{tabular}
         \begin{tabular}{@{\hskip1pt}c@{\hskip1pt}c@{\hskip1pt}c@{\hskip1pt}c@{\hskip1pt}c@{\hskip1pt}c}
            ($e$) & \includegraphics[trim= 40 50 40 70,clip,width=0.3\linewidth]{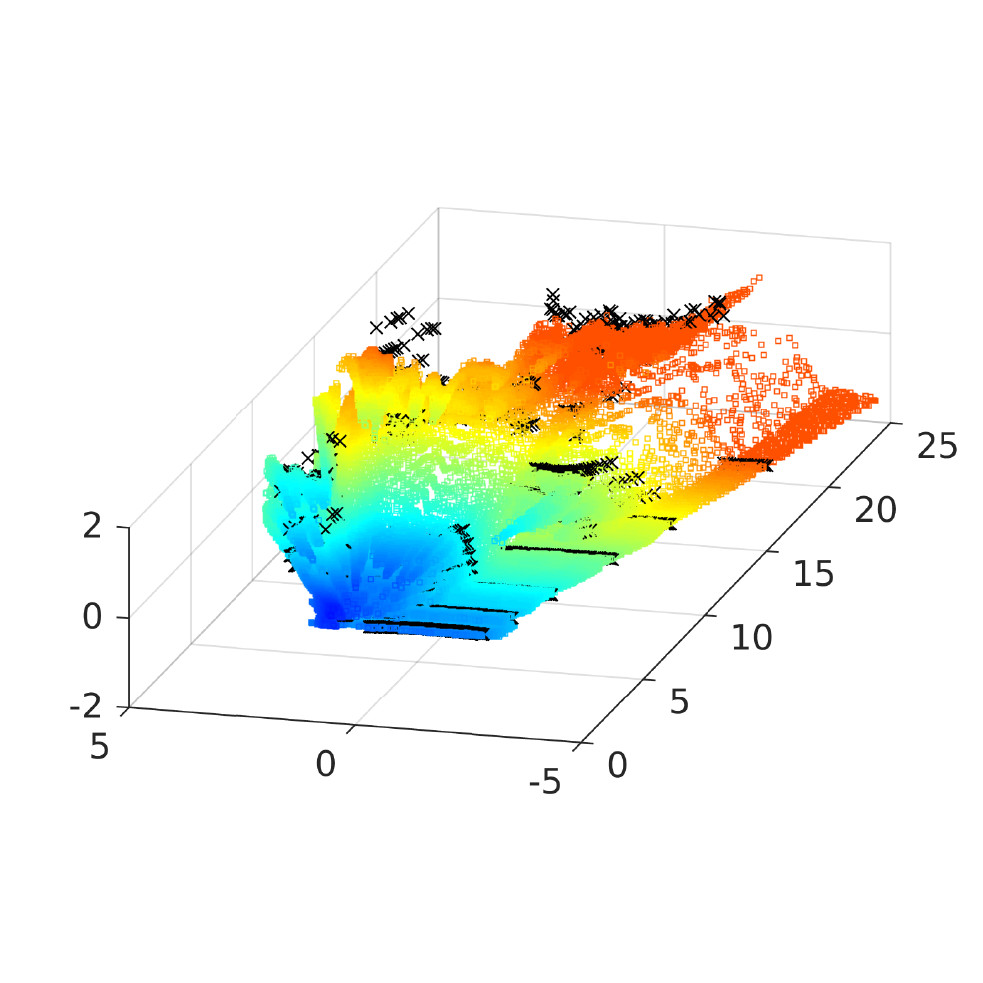} & ($f$) &
            \includegraphics[trim = 40 50 40 70,clip,width=0.3\linewidth]{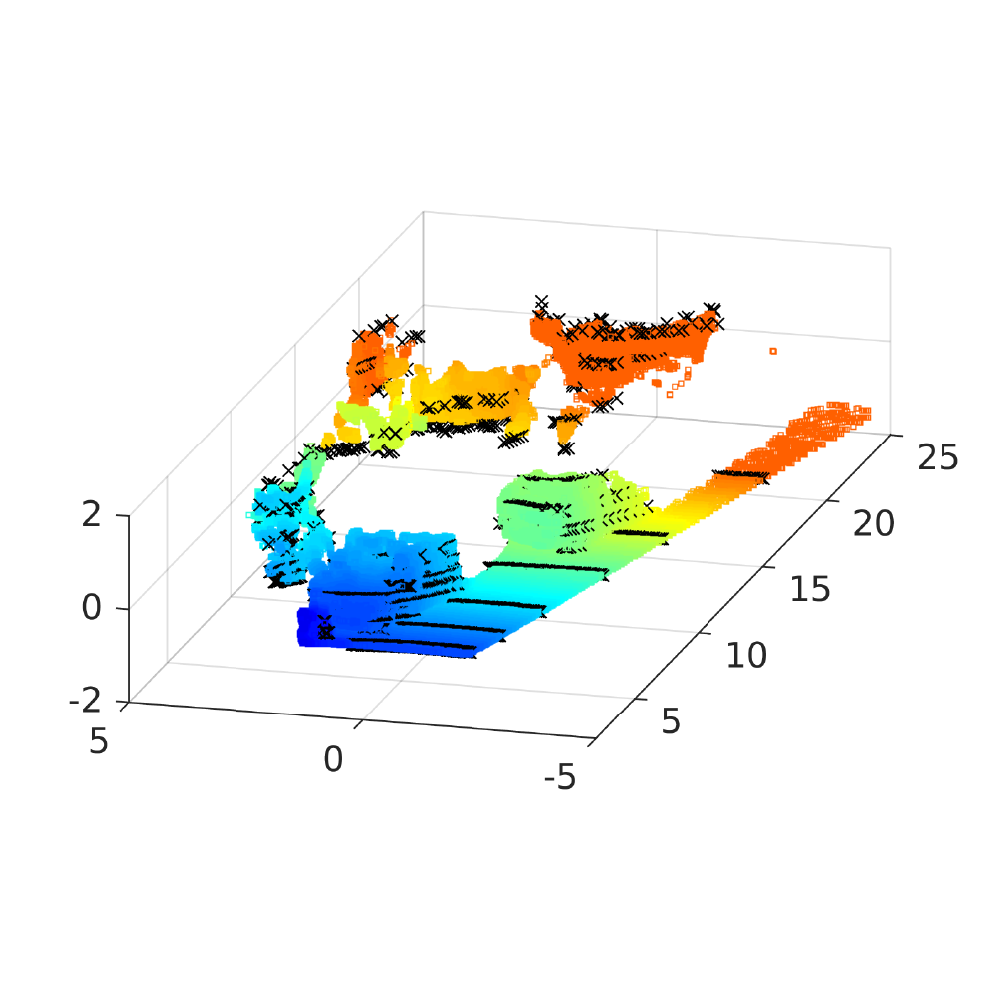} & ($g$) &
            \includegraphics[trim = 40 50 40 70,clip,width=0.3\linewidth]{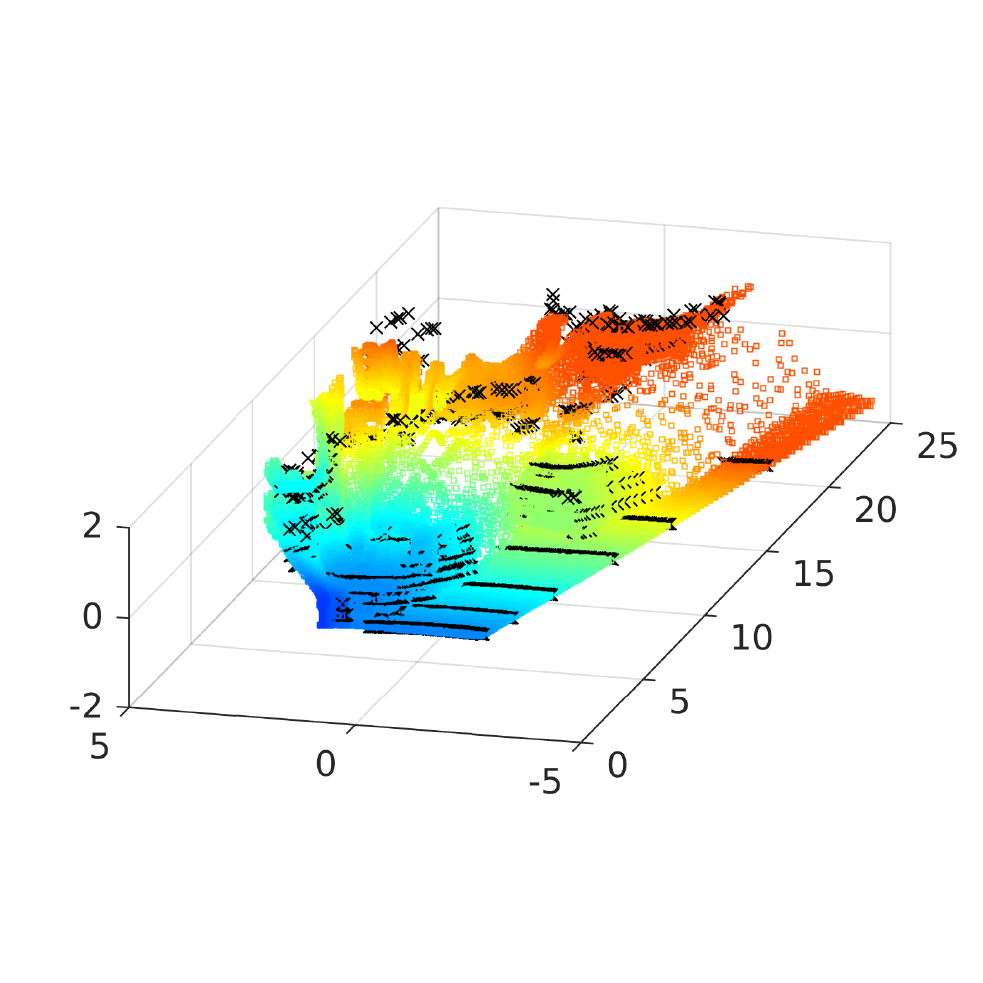}
         \end{tabular}
    \end{center}
    \vspace{-3mm}
    \caption{Another depth completion example with $16$-row Lidar, where all subfigures are defined the same as Fig.~\ref{fig:pedestrian}. Interestingly, higher RMSE is reported on $3$-coefficient estimation as opposed to all-coefficient estimation.}
    \label{fig:car}
\end{figure*}

Depth completion on uniformly subsampling tends to be easier than irregular subsampling; Ma {\em et al.}~\cite{ma2018self} reported improved performance with uniform subsampling.  But in real scenarios, sparse sensors such as Lidar often generate non-uniform, structured patterns when projected into the image plane.  Since our application is to estimate dense depth from inexpensive Lidars in outdoor scenes, we simulate lower resolution Lidars by subsampling $32$ and $16$ rows from $64$R Lidar (depth acquisition sensor used by KITTI). We subsample the points based on selecting a subset of evenly spaced rows of $64$R raw data provided by KITTI (splitted based on the azimuth angle in Lidar space) and then projecting the points into the image.

\vspace{-0.5cm} 
\paragraph{Error Metrics}
We use the standard error metrics: RMSE, MAE, Mean Absolute Relative Error (MRE), and $\delta_i$. $\delta_i$ is the percentage of predicted pixels whose relative error is within a relative threshold (higher being better), defined as:
\begin{align}
    \delta_i : \frac{\mathbf{card}\max({\frac{\hat{y_i}}{y_i},\frac{y_i}{\hat{y_i}})}<\delta_i}{\mathbf{card}( \{y_i \})}
\end{align}    
 We also include our proposed metrics: tMAE, and tRMSE.

\vspace{-0.5cm} 
\paragraph{Implementation Details}
The experiment is implemented in Tensorflow 1.10~\cite{tensorflow2015-whitepaper}. We use Adam optimizer for training with an initial learning rate of $10^{-4}$ and decreased to half every $5$ epochs. We use a single GPU $1080$~Ti with $32$G RAM for training and evaluation. Since GPU memory does not support full-sized KITTI images, we train it with patches of size $224 \times 224$ and batch-size of $3$. For NYU2 dataset, we select batch-size of $10$ and continue the training for $7-10$ epochs.

\subsection{Results}

Table~\ref{tab:nyu2_results} shows a comparison on the NYU2 dataset. Our method shows improvements in all metrics except RMSE.  And interestingly, unlike CSPN + Unet~\cite{cheng2018depth}, our method requires no fine-tuning networks (which increases the inference time) to sharpen boundaries.  

Table~\ref{tab:kitti_results_valid} reports a quantitative comparison of our method with our implementation of Ma {\em et al.}~\cite{ma2018self} performing depth completion on $16$-row Lidar. We select $16$-row Lidar since they are inexpensive and commercially feasible for automakers. Results are on KITTI's validation set of $1,000$ images.  Due to GPU memory constraints, our patch sizes and batch sizes were smaller and so our implementation performance is lower than in~\cite{ma2018self}.   However, from a comparison perspective, our network architecture is the same, see Fig.~\ref{fig:nn_architecture}, and so the improved results of our method are due to using DC input and cross-entropy loss on the output.  

\subsection{Ablation Studies}


Table~\ref{tab:spar_perf} shows depth completion performance as Lidar sparsity increases.  At each sparsity level results are shown for depth estimated from $3$ coefficients, see Eq.~\ref{eq:estDC}, and all coefficients, Eq.~\ref{eq:DCP}.  While they both have roughly the same MAE, $3$-coefficient prediction has smaller tMAE and tRMSE but larger RMSE. Likely this is due to fewer mixed-depth pixels, as can be seen in Figs.~\ref{fig:pedestrian} and \ref{fig:car}.

\begin{table}[t!]
    \centering
    \setlength\doublerulesep{0.5pt}
    \begin{tabular}{|c|c|c|c|c|}
        \hline
        Sparsity & MAE & RMSE & tMAE & tRMSE \\ 
            \hhline{=====} 
          $64$R-$3$coeff & $24.1$ & $121.2$ & $20.3$ & $34.4$ \\
          $64$R-all & $25.2$ & $106.1$ & $23.9$ & $37.4$\\
          $32$R-$3$coeff & $31.0$ & $132.2$ & $24.4$ & $39.5$ \\
          $32$R-all & $31.1$ & $115.8$ & $27.6$ & $42.2$\\
          $16$R-$3$coeff & $37.8$ & $160.6$ & $33.4$ & $47.2$ \\
          $16$R-all & $38.6$ & $142.3$ & $36.1$ & $50.5$\\
          \hline
    \end{tabular}
    \caption{Performance evaluation at different levels of Lidar sparsity (KITTI dataset). Units in cm.}
    \label{tab:spar_perf}
\end{table}
\begin{table}[t!]
    \centering
    \setlength\doublerulesep{0.5pt}
    \begin{tabular}{|c|c|c|c|c|c|}
        \hline
        Input & Loss & MAE & RMSE & tMAE & tRMSE \\ \hhline{======}
          SP & MSE & $6.63$ & $15.28$ & $5.96$ & $6.97$\\
          DC & MSE & $6.10$ & $15.32$ & $5.72$ & $6.73$\\ 
          SP & CE & $9.53$ & $17.81$ & $6.75$ & $7.56$ \\
          DC & CE & \boldsymbol{$3.82$} & \boldsymbol{$11.85$} & \boldsymbol{$4.24$} & \boldsymbol{$5.37$} \\ 
          \hline
    \end{tabular}
    \caption{A comparison whether DC on the input or DC with cross entropy (CE) on output has the dominant effect.  It turns out that individually their effect is small, but together have a large impact (NYU2 dataset). Units in cm.}
    \label{tab:optim_inpout_perf}
\end{table}

One interesting question is whether the gains we are seeing are coming from use of DC on the input or use of DC plus cross-entropy on the output.  Table~\ref{tab:optim_inpout_perf} compares all four combinations of inputs and outputs and finds that by far the biggest gains are when DC is used in both input and output.
We ablate on how the number of DC channels affects efficiency, in Tab.~\ref{tab:inference_kitti}. In each of the variation, we create the DC from $2.5$D depth and recover the $2.5$D depth from DC on-the-fly. There is some computational penalty to DC, but it is relatively small, and can be remedied by reducing the number of channels.

Another application of depth completion is to improve on object detection.  While it might seem intuitive that at higher resolution, estimated dense depth could give better vehicle detection, often this is not the case, and we are not aware of other past literature reporting this.  Likely mixed-depth pixels have a large negative impact on object detection.  Indeed, Tab.~\ref{tab:detect_obj} shows worse car detection on Ma's output than on the raw $16$-row sparse data.  However, our method is able to outperform sparse depth, an important step towards improving Lidar-based object detection.

\begin{table}[t!]
    \centering
    \setlength\doublerulesep{0.5pt}
    \resizebox {\linewidth}{!}{
    \begin{tabular}{|c|c|c|c|}
    \hline
        Method & MAE (cm) & RMSE (cm) & Infer.~time (ms) \\
        \hhline{====}
        $2.5$D (SP)-$1$C. & $65.2$ & $174$ & $130$ \\ 
        DC-$10$C-$3$coeff & $50.2$ & $169$ & $140$ \\
        DC-$20$C-$3$coeff & $45.4$ & $165$ & $145$ \\ 
        DC-$40$C-$3$coeff & $37.8$ & $161$ & $161$ \\
        DC-$80$C-$3$coeff & $38.4$ & $167$ & $202$ \\
        \hline
          
    \end{tabular}
    }
    \caption{Performance and efficiency on KITTI validation benchmark using $16$R Lidar points.} 
    \label{tab:inference_kitti}
    \vspace{-3mm}
\end{table}

\begin{table}[t!]
	\begin{center}
    \scalebox{0.9}{%
		\begin{tabular}{|c|c|c|c|c|c|c|}
			\hline
			&\multicolumn{3}{|c|}{3D Bounding Box} &\multicolumn{3}{|c|}{Bird's Eye View Box} \\
			\cline{2-7} 
			Upsample: & Easy & Med. & Hard & Easy & Med. & Hard \\
			\hhline{=======}
			Raw $16$R & $54.4$ & $36.2$ & $31.3$ & $73.6$ & \boldsymbol{$58.1$} & \boldsymbol{$50.4$}\\
			Ma~\cite{ma2018self} & $36.7$ & $23.0$ & $18.5$ & $56.2$ & $33.8$ & $29.7$\\
			DC-$3$coeff & \boldsymbol{$64.9$} & \boldsymbol{$41.9$} & \boldsymbol{$34.7$} & \boldsymbol{$78.1$} & $54.0$ & $45.6$\\
			\hline
		\end{tabular}}
		\end{center}
		 \vspace{-2mm}
	    \caption{Average precision (\%) for $3$D detection and pose estimation of cars on KITTI~\cite{Geiger2012CVPR} using Frustum PointNet~\cite{qi2017frustum}. The baseline, Raw-$16$R, uses $16$ rows from the Lidar, while Ma's method~\cite{ma2018self} and our method start by densely upsampling these $16$R data.  In each case, the method is trained on $3,712$ frames and evaluated on $3,769$ frames, of the KITTI $3$D object detection benchmark~\cite{Geiger2012CVPR} using an intersection of union (IOU) measure of $0.7$.  Only our method improves on the baseline, and this is the most significant for $3$D bounding boxes.}
		\label{tab:detect_obj}
\end{table}

\section{Conclusion}

Upsampling depth in a manner that respects object boundaries is challenging.  Deep networks have shown progress in achieving this, but nevertheless still generate mixed-depth pixels.  Our work tackles this problem on both the input and the output sides of a prediction network.  On the input, our Depth Coefficients represent depth without loss in accuracy (unlike binning) while separating pixels by depth so that it is simple for convolutions to avoid depth mixing.  On the output side, instead of directly predicting depth, we predict a depth density using cross entropy on the Depth Coefficients.  This is a richer representation that avoids depth mixing and can enable deeper levels of fusion and object detection.  Indeed we show that, unlike other upsampling methods, our dense depth estimates can improve object detection compared to sparse depth.  Including Depth Coefficients on the input and output of networks is an easy and simple way to achieve better performance.

We showed that in the case of ambiguities, MSE is a flawed metric to evaluate depth completion.  Now that depth completion methods are producing high-quality dense depths, our proposed metrics, tRMSE and tMAE, are preferable as they reward high-probable depth estimates and give equal penalty to large errors, which are mostly mixed-depth pixels.

\section*{Acknowledgements}
This work was partially supported with funds from Changan US, and the authors greatly appreciate inputs from Radovan Miucic, Ashish Sheikh and Gerti Tuzi.

{\small
\bibliographystyle{ieee}
\bibliography{abbrev,egbib}
}

\end{document}